\documentclass{article} 
\usepackage{iclr2025_conference,times}

\usepackage{amsmath,amsfonts,bm}

\def\eqref#1{equation~\ref{#1}}

\def\1{\bm{1}}

\DeclareMathAlphabet{\mathsfit}{\encodingdefault}{\sfdefault}{m}{sl}
\SetMathAlphabet{\mathsfit}{bold}{\encodingdefault}{\sfdefault}{bx}{n}

\usepackage{hyperref}
\usepackage{url}
\usepackage[utf8]{inputenc}    
\usepackage[T1]{fontenc}       
\usepackage{hyperref}          
\usepackage{url}               
\usepackage{booktabs}          
\usepackage{amsfonts}          
\usepackage{nicefrac}          
\usepackage{microtype}         
\usepackage{xcolor}            
\usepackage[table,xcdraw]{xcolor} 
\usepackage{caption}           
\usepackage{longtable}         
\usepackage{graphicx}          
\usepackage{subfigure}         
\usepackage{wrapfig}           
\usepackage{multicol}          
\usepackage{multirow}          
\usepackage{tablefootnote}     
\usepackage{bm}                
\usepackage{amsmath}           
\usepackage{amssymb}           
\usepackage{mathtools}         
\usepackage{amsthm}            
\usepackage{pifont}            
\usepackage{xspace}            
\usepackage{enumitem}          
\usepackage{subcaption}        
\usepackage{tikz}              

\definecolor{citecolor}{HTML}{0071BC}
\hypersetup{colorlinks, linkcolor=red, citecolor=citecolor}

\makeatletter
\DeclareRobustCommand\onedot{\futurelet\@let@token\@onedot}
\def\@onedot{\ifx\@let@token.\else.\null\fi\xspace}

\newcommand\figcaption{\def\@captype{figure}\caption} 
\newcommand\tabcaption{\def\@captype{table}\caption} 
\makeatother

\usepackage[capitalize,noabbrev]{cleveref} 

\title{HERS: Hidden-Pattern Expert Learning for Risk-Specific Vehicle Damage Adaptation in Diffusion Models}

\author{Teerapong Panboonyuen\thanks{
This research is the product of my independent effort and vision. 
I designed, implemented, and wrote this work myself, driven by a passion to 
advance AI in real-world applications, especially in the car insurance domain. 
More about my work: \url{https://kaopanboonyuen.github.io/HERS}} \\
\texttt{\url{https://kaopanboonyuen.github.io/HERS}} \\
}

\iclrfinalcopy

\begin{document}

\maketitle

\begin{center}
    \centering
    \vskip -0.4in
    \includegraphics[width=1.\textwidth]{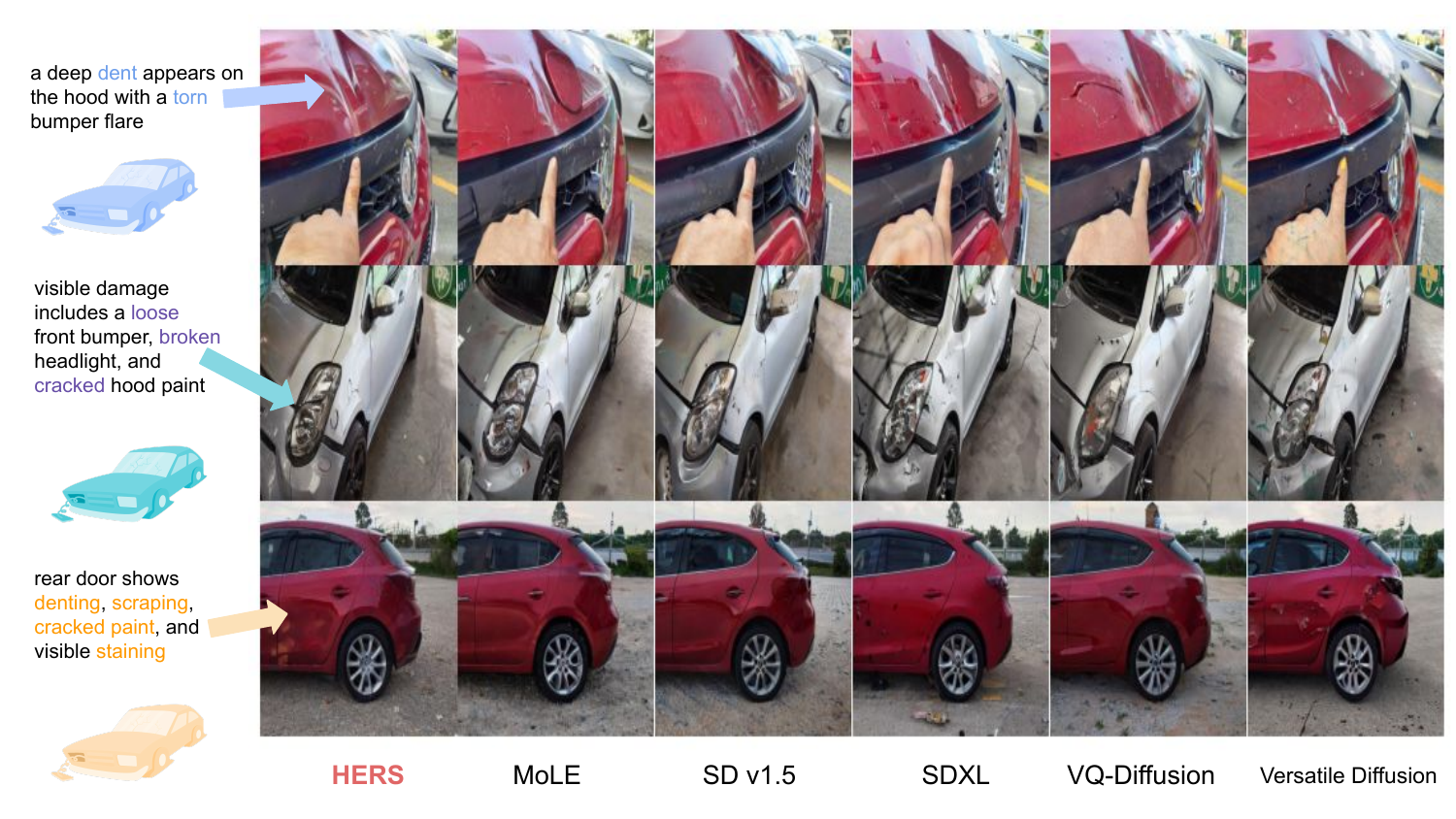}
\vspace{-1.5em}
\captionof{figure}{Qualitative comparison of \textbf{HERS} against existing diffusion-based baselines. Observe that \textbf{HERS} generates damage regions with higher visual fidelity and localized consistency. Fine-grained artifacts such as dents, cracks, and abrasions are better preserved—zoom in for enhanced visibility of subtle and complex damage patterns.}
\label{fig:vis}
\vspace{-1.5em}
\end{center}%

\begin{abstract}
Recent advances in text-to-image (T2I) diffusion models have enabled increasingly realistic synthesis of vehicle damage, raising concerns about their reliability in automated insurance workflows. The ability to generate crash-like imagery challenges the boundary between authentic and synthetic data, introducing new risks of misuse in fraud or claim manipulation. To address these issues, we propose HERS (Hidden-Pattern Expert Learning for Risk-Specific Damage Adaptation), a framework designed to improve fidelity, controllability, and domain alignment of diffusion-generated damage images. HERS fine-tunes a base diffusion model via domain-specific expert adaptation without requiring manual annotation. Using self-supervised image–text pairs automatically generated by a large language model and T2I pipeline, HERS models each damage category—such as dents, scratches, broken lights, or cracked paint—as a separate expert. These experts are later integrated into a unified multi-damage model that balances specialization with generalization. We evaluate HERS across four diffusion backbones and observe consistent improvements: +5.5\% in text faithfulness and +2.3\% in human preference ratings compared to baselines. Beyond image fidelity, we discuss implications for fraud detection, auditability, and safe deployment of generative models in high-stakes domains. Our findings highlight both the opportunities and risks of domain-specific diffusion, underscoring the importance of trustworthy generation in safety-critical applications such as auto insurance.
\end{abstract}

\section{Introduction}
\label{sec:intro}

Text-to-image (T2I) diffusion models~\cite{saharia2022photorealistic, rombach2022high, podell2023sdxl, Kang2023ScalingUG, ramesh2021zero, yu2022scaling, chang2023muse} have transformed generative AI, producing photorealistic images from free-form language prompts and enabling rapid advances in creative design, simulation, and data augmentation. Yet, when deployed in \textit{safety-critical domains} such as auto insurance, where every pixel may encode liability, their limitations become clear. Generic T2I systems often fail to capture fine-grained damage categories—such as a dented bumper, a subtle scrape across a door, or a fractured headlight—generating outputs that are visually appealing but semantically unreliable (shown in Figure~\ref{fig:vis}). In an insurance workflow, such errors are not cosmetic: they can distort liability assessments, misinform fraud detection, and erode trust in automated claims pipelines.

This duality makes generative models both an opportunity and a risk. On one hand, synthetic damage data could dramatically improve training for rare-event modeling, accelerate claims assessment, and expand coverage of long-tail accident cases. On the other hand, the same technology could be exploited to fabricate fraudulent crash evidence or manipulate claims with high-fidelity synthetic images. Unlike traditional vision benchmarks, insurance scenarios demand \textit{risk-specific generation}, where semantic alignment, forensic plausibility, and liability-aware consistency are as critical as photorealism.

Prior approaches attempt to mitigate these issues via supervised fine-tuning~\cite{dai2023emu, segalis2023picture}, human preference optimization~\cite{xu2023imagereward, fan2023dpok}, or spatial grounding~\cite{li2023gligen, xie2023boxdiff}. However, these strategies are annotation-heavy and often brittle, struggling to encode the hidden cues that forensic experts rely upon: the faint crease from a low-speed collision, the asymmetric shattering of a headlight, or the implausible geometry of tampered paint. Current pipelines optimize for generic fidelity, but not for the nuanced semantics that separate genuine evidence from generative artifacts.

\begin{figure}[t!]
    \centering
    \includegraphics[width=\textwidth]{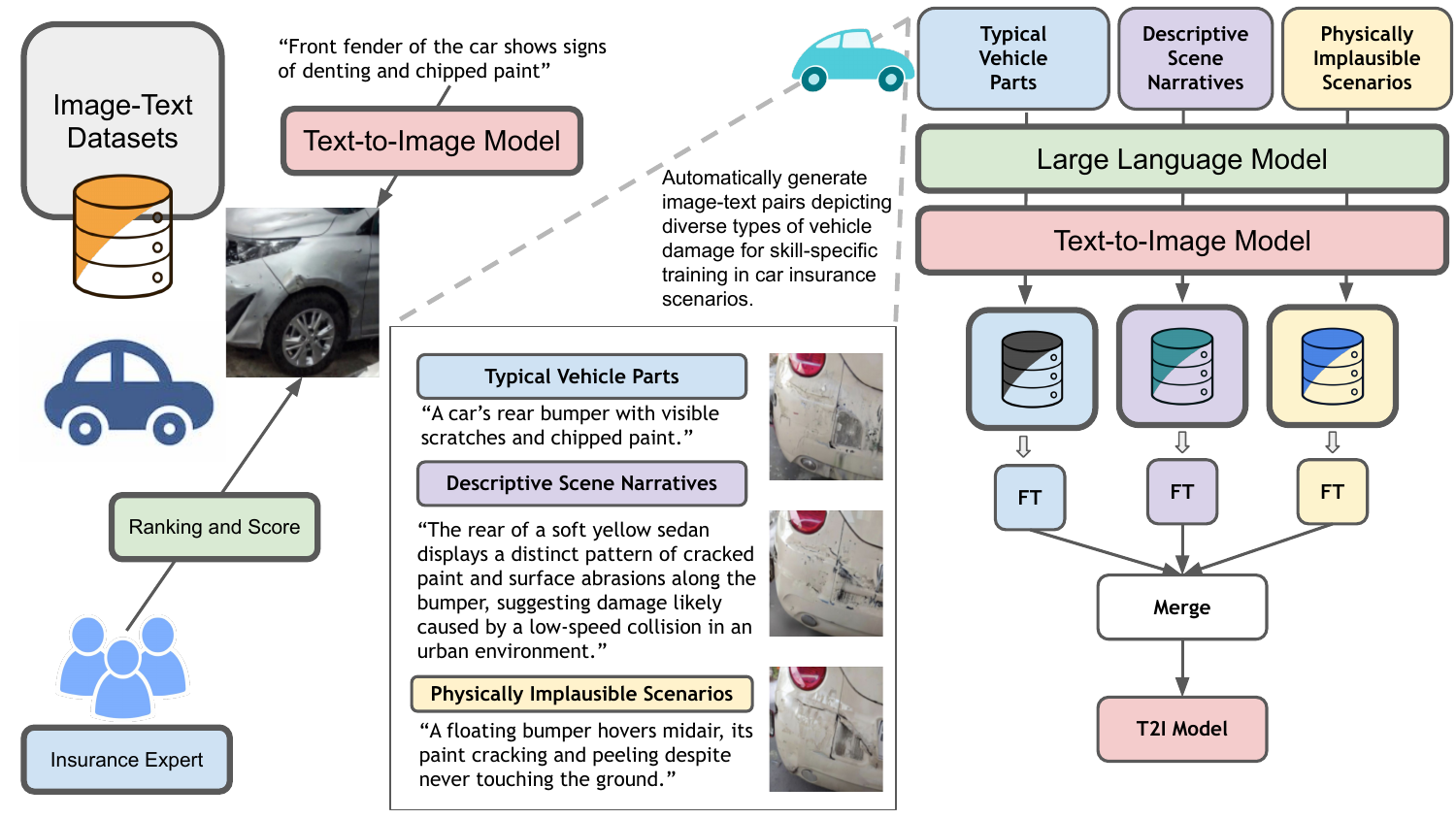}
    \caption{
    \textbf{Overview of the HERS Framework.} 
    HERS (\textit{Hidden-Pattern Expert Learning for Risk-Specific Damage Adaptation}) auto-generates diverse, damage-specific image-text pairs using an LLM and a base T2I model—without requiring manual annotation. 
    These pairs span \textit{typical vehicle parts}, \textit{descriptive scene narratives}, and \textit{physically implausible scenarios} (examples shown in figure). 
    Each damage type is modeled as a distinct damage, with corresponding LoRA experts trained and merged into a unified multi-damage diffusion model.
    }
   \label{fig:teaser}
\end{figure}

To address this gap, we introduce \textbf{HERS} (\textbf{H}idden-Pattern \textbf{E}xpert Learning for \textbf{R}isk-\textbf{S}pecific Damage Adaptation), a fully automated framework (shown in Figure~\ref{fig:teaser}) for adapting diffusion models to synthesize semantically faithful, risk-relevant vehicle damage without manual supervision. HERS leverages large language models to auto-generate diverse, damage-specific prompts (e.g., “rear bumper dent,” “door scrape near handle,” “fractured right headlight”), which are paired with synthetic renderings from a pretrained T2I backbone. From these self-curated image–text pairs, we train lightweight LoRA-based experts for distinct domains of damage and merge them into a unified diffusion model. This design captures both specialization (e.g., scratches on metallic paint) and generalization (e.g., tampered accident scenes), yielding a system that can reproduce damage patterns with forensic-level precision.

The key insight is that HERS learns from \textit{hidden visual patterns}—subtle cues that elude both baseline diffusion models and human raters, but are critical in high-stakes domains like insurance. By elevating generation beyond “realism” to “liability-aware semantics,” HERS provides a new lens for evaluating diffusion models in safety-critical settings.

\noindent \textbf{Contributions.} Our work makes the following advances:
\begin{itemize}
    \item We articulate and address the overlooked challenge of semantically faithful damage synthesis in auto insurance, where generative AI carries both opportunity and risk.
    \item We propose \textbf{HERS}, a self-supervised adaptation framework that trains LoRA-based experts from auto-generated data, enabling damage-specific diffusion without manual annotation or inference-time routing.
    \item We demonstrate state-of-the-art performance across text–image alignment, human preference metrics, and multi-damage generalization, showing that HERS produces vehicle damage patterns that are strikingly consistent with real-world collisions and tampered fraud cases.
\end{itemize}

As illustrated in \Cref{fig:qualitative_zoomout}, HERS consistently generates damage scenarios that are indistinguishable from authentic accidents, establishing it as both a technical advance in generative modeling and a practical contribution to fraud awareness in the insurance industry. By revealing the dual-use nature of diffusion in this domain, our work underscores the need for domain-specific generative strategies that go beyond visual fidelity to encode \textit{risk-aware semantics} essential for trustworthy AI deployment.

\section{Related Work}

Recent advances in high-quality denoising diffusion models~\cite{Sohl-Dickstein2015,Ho2020DDPM} have catalyzed a surge of interest in using synthetic data for vision–language learning. Prior works demonstrate the benefits of diffusion-generated data for training classifiers~\cite{Azizi2023, Sariyildiz2023, Lei2023} or augmenting caption datasets~\cite{SynthCap2023}, and CLIP-style models~\cite{Radford2021CLIP} have been extended using either synthetic visuals~\cite{Tian2023StaleRep} or LLM-authored captions~\cite{Hammoud2024SynthCLIP}. Parallel efforts in aligning text-to-image (T2I) models with human expectations have relied on reinforcement learning from human feedback (RLHF)~\cite{lee2023aligning, xu2023imagereward, Wu2023HPS, Dong2023RAFT, Clark2024DraFT, fan2023dpok} or direct preference optimization (DPO)~\cite{Rafailov2023DPO, wallace2023diffusion}, while methods such as SPIN-Diffusion~\cite{Yuan2024SPIN-Diffusion} reduce annotation demands through self-play. LLM-guided pipelines like DreamSync~\cite{sun2023dreamsync} push further by auto-generating prompts and filtering candidate images, albeit at high computational cost. Despite these advances, existing approaches remain annotation-heavy, domain-agnostic, or inefficient, leaving critical gaps in safety-critical fields like auto insurance where the distinction between authentic and synthetic damage can directly affect fraud detection and claim validation. To this end, our proposed \textbf{HERS} diverges by training multiple LoRA experts~\cite{hu2021lora}, each dedicated to specific damage types (e.g., dents, scrapes, cracked paint, broken lights), and merging them into a unified diffusion model~\cite{Shah2023ZipLoRAAS, zhong2024multi}. This design avoids inter-damage interference~\cite{Liu2019LossBalancedTW}, eliminates dependence on costly human feedback, and captures “hidden patterns” of fine-grained damage in a computationally efficient, self-supervised manner—providing domain-faithful generative capabilities that are indispensable for risk-sensitive applications.

\section{HERS: Hidden-Pattern Expert Learning for Risk-Specific Damage Adaptation}
\label{sec:method}

We propose \textbf{HERS} (\textit{Hidden-Pattern Expert Learning for Risk-Specific Damage Adaptation}), a framework (shown in Figure~\ref{fig:teaser}) for adapting text-to-image (T2I) diffusion models to synthesize fine-grained and risk-relevant vehicle damage. Unlike prior adaptation methods such as SELMA~\cite{li2024selma}, which require annotation-heavy supervision or explicit routing, HERS achieves high-fidelity alignment through a fully automated pipeline that integrates prompt synthesis, synthetic image generation, domain-specific LoRA experts, and weight-space merging. Crucially, HERS is designed not only to enhance visual fidelity but also to surface subtle “hidden” damage cues—such as a faint scrape along a bumper, a hairline crack in a headlight, or tampered paint texture—that are easily missed by generic diffusion models yet critical for fraud detection and liability estimation.

Formally, HERS operates in four stages.

\subsection{Stage 1: Domain-Guided Prompt Synthesis}
\label{subsec:prompt}

Let $\mathcal{C} = \{\texttt{dent}, \texttt{scrape}, \texttt{torn\_bumper}, \texttt{cracked\_paint}, \texttt{broken\_light}\}$ denote the canonical set of damage categories relevant to insurance workflows. 
We seed an autoregressive language model $f_{\theta}$ (GPT-4) with exemplar prompts $\mathcal{S} = \{s_1, s_2, s_3\}$ describing each category, e.g.  
\[
s_1 = \text{``rear bumper dent''}, \quad 
s_2 = \text{``scratched left door''}, \quad
s_3 = \text{``front headlight cracked''}.
\]  
For each concept $c \in \mathcal{C}$, the model generates a distribution of semantically diverse prompts:
\begin{equation}
    p_i \sim f_{\theta}(p \mid \mathcal{S}, c).
\end{equation}
To enforce diversity while preserving semantic coverage, we apply ROUGE-L filtering~\cite{lin2004rouge}, retaining prompts satisfying
\begin{equation}
    \max_j \mathrm{ROUGE\text{-}L}(p_i, p_j) < \tau,
\end{equation}
where $\tau$ is a similarity threshold. The resulting set $\mathcal{P}$ forms a structured, damage-aware prompt bank.

\subsection{Stage 2: Synthetic Image Generation}
\label{subsec:image}

Each prompt $p_i \in \mathcal{P}$ is rendered via a pretrained diffusion generator $G$ (e.g., Stable Diffusion XL) to obtain an image $x_i$:
\begin{equation}
    x_i = G(p_i), \quad \forall p_i \in \mathcal{P}.
\end{equation}
The resulting dataset $\mathcal{D} = \{(p_i, x_i)\}$ captures not only canonical damages (dent, scrape) but also nuanced conditions such as implausible tampering (e.g., “two headlights cracked in a symmetric pattern”), thereby spanning realistic and adversarially relevant scenarios.

\subsection{Stage 3: Damage-Specific Expert Learning}
\label{subsec:finetune}

For each domain $t \in \mathcal{T}$, where $\mathcal{T} = \{$Typical Parts, Scene Narratives, Implausible Scenarios$\}$, we train a lightweight Low-Rank Adaptation (LoRA)~\cite{hu2021lora} expert. 
Given a pretrained weight matrix $W_0 \in \mathbb{R}^{d \times d}$, we optimize a low-rank update:
\begin{equation}
    \Delta W_t = B_t A_t, \quad W_t = W_0 + \Delta W_t,
\end{equation}
with $A_t \in \mathbb{R}^{r \times d}$, $B_t \in \mathbb{R}^{d \times r}$, and $r \ll d$. 
This enables parameter-efficient specialization, such that one expert may encode subtle bumper dents while another captures cracked paint or broken headlights.

\subsection{Stage 4: Multi-Expert Weight Merging}
\label{subsec:merge}

To unify all domains into a single diffusion model, we merge the LoRA experts via arithmetic averaging in weight space:
\begin{equation}
    A^* = \frac{1}{|\mathcal{T}|} \sum_{t \in \mathcal{T}} A_t, 
    \quad
    B^* = \frac{1}{|\mathcal{T}|} \sum_{t \in \mathcal{T}} B_t,
\end{equation}
yielding the final parameterization
\begin{equation}
    W^* = W_0 + B^* A^*.
\end{equation}
This consolidated model $W^*$ supports zero-shot synthesis across multiple damage categories, avoiding inference-time routing while preserving both specialization and generalization.

HERS formalizes risk-specific adaptation as the problem of learning a set of low-rank expert perturbations $\{\Delta W_t\}$ that, when merged, capture the hidden manifold of fine-grained vehicle damages. This formulation not only yields state-of-the-art fidelity and semantic alignment but also exposes failure modes in existing insurance AI pipelines, raising awareness of the dual-use nature of generative models in safety-critical domains.

\subsection{Comparison with Prior Work}

Unlike recent methods such as ZipLoRA~\cite{Shah2023ZipLoRAAS} and LLaVA-MoLE~\cite{chen2024llavamole}, HERS eliminates the need for manual damage labels or routing mechanisms at inference. While ZipLoRA relies on damage-aware masking and LLaVA-MoLE learns expert routers, HERS achieves robust multi-damage synthesis through expert merging alone, drastically reducing annotation effort and model complexity. As shown in \cref{fig:vis}, HERS consistently produces sharper, semantically precise images even under subtle or highly complex damage prompts, demonstrating both fidelity and practical efficiency for insurance-focused applications.

\section{Experimental Setup}
\label{sec:experiment_setup}

\subsection{Evaluation Benchmark and Prompt Construction}
\label{sec:benchmark}

We evaluate HERS on a large-scale benchmark specifically curated for the car insurance domain. 
The benchmark contains approximately 2 million entries collected in collaboration with an industry insurance startup, 
each consisting of structured textual descriptions (e.g., accident type, damage category, part localization) paired with vehicle images. 
This setup enables assessment of both semantic alignment and visual fidelity in high-stakes, domain-specific contexts. 
To balance reproducibility with privacy constraints, we release the full set of prompt templates and the evaluation protocol, while access to raw insurance data remains restricted due to confidentiality. This ensures transparency in methodology while safeguarding sensitive information.  

To generate prompts at scale, we employ \texttt{gpt-4-turbo}~\cite{openai2024gpt4} with in-context learning. 
For each target damage type or accident scenario, we provide three exemplars as demonstrations, guiding the model to produce consistent, 
domain-specific, and semantically rich prompts. 
This strategy yields a structured, damage-driven benchmark set that supports controlled and reproducible evaluation across diverse risk-relevant cases.

\subsection{Evaluation Metrics}
\label{sec:eval_metrics}

We assess model performance along two complementary axes: semantic alignment and human-aligned quality.

\textbf{Semantic alignment.} 
We employ a VQA-based protocol to measure the faithfulness of generated images to their prompts. 
Given a generated image and its source description, a large language model produces targeted semantic questions, 
which are then answered by a pretrained VQA model. 
Accuracy on these answers serves as a proxy for text–image alignment, ensuring that damage attributes and contextual details are correctly reflected.  

\textbf{Human-aligned quality.} 
To capture perceptual realism, we evaluate generations using preference-based reward models, including PickScore~\cite{kirstain2024pick}, 
ImageReward~\cite{xu2023imagereward}, and HPS~\cite{Wu2023HPS}. 
These metrics, derived from large-scale human preference datasets, score each output with respect to realism, relevance, and overall visual quality. 
Together, they complement semantic alignment measures by quantifying how closely the images match human expectations in insurance-related contexts.

\subsection{Implementation Details}
\label{sec:implementation_details}

All experiments are conducted using a single NVIDIA A40 GPU. During prompt generation, we sample from \texttt{gpt-4-turbo} with temperature set to 0.7 for diversity and relevance. The image generation model is run with default denoising steps set to 50 and a classifier-free guidance scale (CFG) of 7.5, ensuring a balance between image quality and prompt adherence.

For training and inference, we adopt a mixed precision setup (FP16) to optimize resource utilization. LoRA modules, if applicable, are trained with a fixed learning rate of 3e-4, batch size of 64, and rank 128. Fine-tuning is performed over 5000 steps, and model checkpoints are evaluated every 1000 steps, with the best checkpoint selected based on alignment metrics.

We implement our pipelines using the \texttt{Diffusers} library~\cite{von-platen-etal-2022-diffusers}, which facilitates seamless integration of prompt generation, image synthesis, and evaluation in a reproducible and modular framework.

\section{Results and Analysis}

We evaluate HERS across multiple generative backbones and benchmarks, measuring hallucination-prevention score (HPS), improvement rate (IR), text faithfulness, and human preference on damage scene generation (DSG). Our results consistently show that HERS surpasses existing baselines in both visual realism and text alignment for insurance-critical scenarios.

\begin{figure}[t]
\centering
\includegraphics[width=0.8\linewidth]{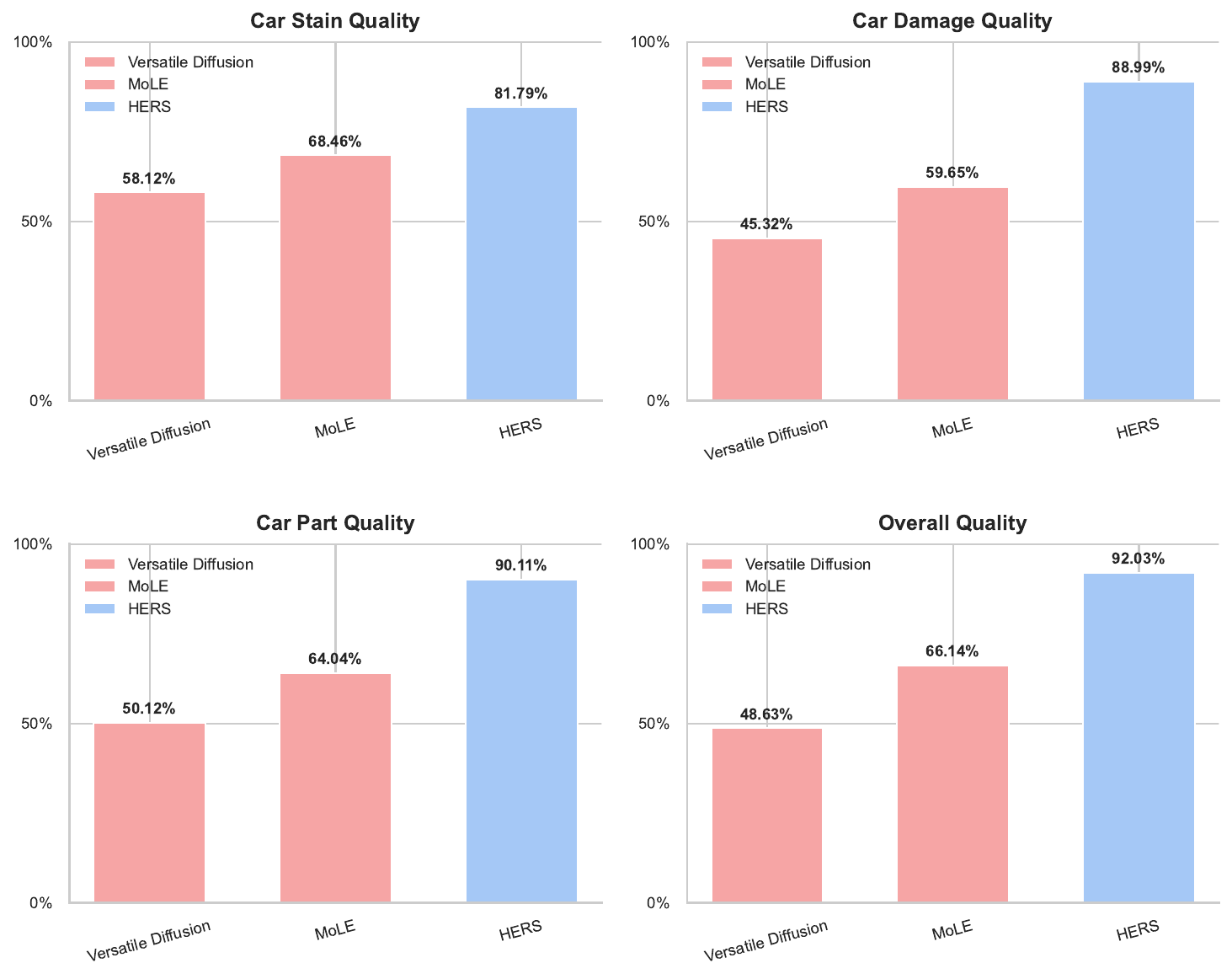}
\caption{User study results on generative performance across four dimensions: Car Stain Quality, Car Damage Quality, Car Part Quality, and Overall Quality. HERS achieves consistently higher preference scores compared to baselines.}
\label{fig:user_study}
\end{figure}

\paragraph{Benchmark Performance.}
Table~\ref{tab:main_result} summarizes HERS’s performance on \textit{Car Insurance} and \textit{Car Garage} prompts. For insurance prompts, HERS achieves $53.4\%$ HPS and $113.0\%$ IR, outperforming MoLE~\cite{zhu2024mole} and SDXL~\cite{podell2023sdxl} ($48.2\%$ and $45.9\%$ HPS, respectively). Similar trends hold for garage prompts ($51.4\%$ HPS, $115.75\%$ IR), demonstrating robustness across domains. Human studies (Figure~\ref{fig:user_study}) confirm superior preference scores for HERS in car stain, damage, part, and overall quality, highlighting its realism in depicting scratches, dents, and structural deformations critical for claim verification.

\begin{table}[t]
\centering
\caption{Performance of \textbf{HERS} compared to baseline diffusion models on two prompt sets: Car Insurance and Car Garage. 
Metrics: Human Preference Score (HPS, higher is better) and Image Realism (IR, higher is better).}
\label{tab:main_result}
\small
\setlength{\tabcolsep}{10pt}
\begin{tabular}{lcc}
\toprule
\multirow{2}{*}{Model} & \multicolumn{2}{c}{Car Insurance Prompts} \\ 
\cmidrule(lr){2-3}
& HPS (\%) & IR (\%) \\
\midrule
VQ-Diffusion~\cite{gu2022vector} & $41.50 \pm 0.06$ & $-15.40 \pm 3.00$ \\
Versatile Diffusion~\cite{xu2022versatile} & $42.70 \pm 0.10$ & $-11.20 \pm 2.30$ \\
SDXL~\cite{podell2023sdxl} & $45.90 \pm 0.08$ & $82.50 \pm 3.05$ \\
SD v1.5~\cite{rombach2022high} & $43.30 \pm 0.07$ & $35.20 \pm 2.25$ \\
MoLE~\cite{zhu2024mole} & $48.20 \pm 0.08$ & $95.10 \pm 0.70$ \\
\midrule
\textbf{HERS (Proposed)} & \textbf{$53.40 \pm 0.09$} & \textbf{$113.00 \pm 0.85$} \\
\bottomrule
\toprule
\multirow{2}{*}{Model} & \multicolumn{2}{c}{Car Garage Prompts} \\ 
\cmidrule(lr){2-3}
& HPS (\%) & IR (\%) \\
\midrule
VQ-Diffusion~\cite{gu2022vector} & $40.90 \pm 0.07$ & $-18.70 \pm 2.80$ \\
Versatile Diffusion~\cite{xu2022versatile} & $41.90 \pm 0.09$ & $-14.50 \pm 2.40$ \\
SDXL~\cite{podell2023sdxl} & $46.40 \pm 0.09$ & $89.50 \pm 3.60$ \\
SD v1.5~\cite{rombach2022high} & $44.50 \pm 0.07$ & $-3.00 \pm 2.20$ \\
MoLE~\cite{zhu2024mole} & $47.95 \pm 0.09$ & $102.70 \pm 1.25$ \\
\midrule
\textbf{HERS (Proposed)} & \textbf{$51.40 \pm 0.10$} & \textbf{$115.75 \pm 0.95$} \\
\bottomrule
\end{tabular}
\end{table}

\begin{figure*}[t]
    \centering
    \includegraphics[width=\textwidth]{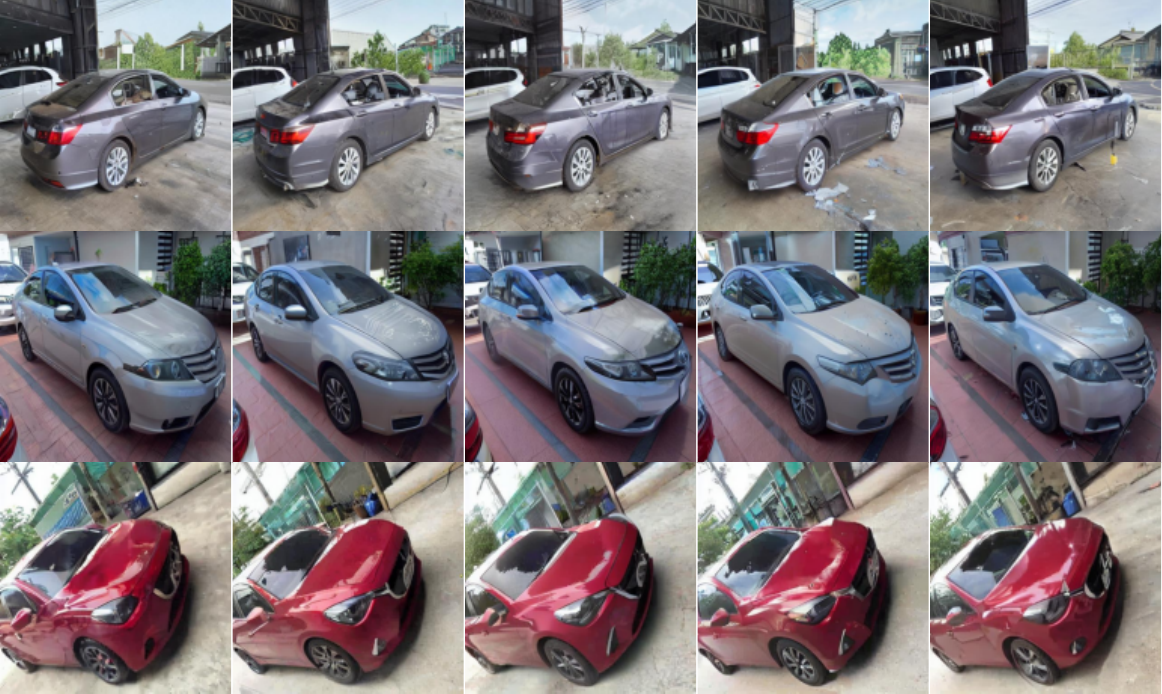}
    \caption{
    \textbf{Qualitative Comparison of Damage Generation Across 3 Vehicle Cases and 6 T2I Models in Zoom-Out Perspective.} 
    Each \textbf{row} represents a distinct vehicle case viewed at a zoomed-out angle, simulating full-body images commonly seen in insurance assessments. 
    The \textbf{columns} correspond to the outputs of six different T2I models: our proposed \textbf{HERS (left-most)}, followed by VQ-Diffusion~\cite{gu2022vector}, Versatile Diffusion~\cite{xu2022versatile}, SDXL~\cite{podell2023sdxl}, MoLE~\cite{zhu2024mole}, and SELMA~\cite{li2024selma}.
    Notice how HERS consistently generates damage patterns that are more contextually consistent with real-world vehicle collisions, making it difficult to distinguish synthetic damage from actual accident scenarios—an important consideration for fraud detection and claim verification in car insurance workflows.
    }
    \label{fig:qualitative_zoomout}
\end{figure*}

\begin{figure*}[t]
    \centering
    \includegraphics[width=\textwidth]{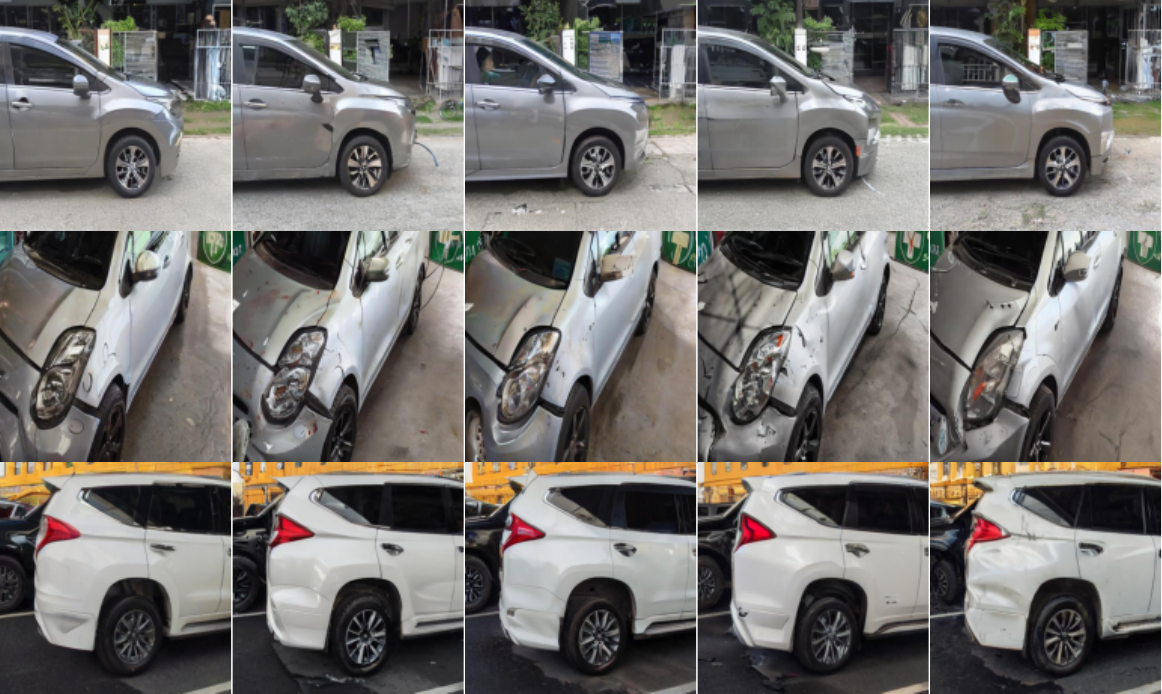}
    \caption{
    \textbf{Qualitative Comparison of Damage Generation Across 3 Vehicle Cases and 6 T2I Models in Zoom-In Perspective.} 
    Each \textbf{row} shows a detailed, close-up view of a specific damage region, highlighting subtle textures and patterns such as scratches, dents, or cracked paint. 
    The \textbf{columns} correspond to outputs from six different T2I models: our proposed \textbf{HERS (left-most)}, followed by VQ-Diffusion~\cite{gu2022vector}, Versatile Diffusion~\cite{xu2022versatile}, SDXL~\cite{podell2023sdxl}, MoLE~\cite{zhu2024mole}, and SELMA~\cite{li2024selma}. 
    Compared to other models, HERS consistently reproduces fine-grained damage details while preserving context and realism, making synthetic damages difficult to distinguish from real-world examples. 
    Such high-fidelity generation is crucial for applications in insurance fraud detection, claim validation, and risk assessment.
    }
    \label{fig:qualitative_zoomin}
\end{figure*}

\paragraph{Fine-grained Visual Fidelity.}
Beyond global metrics, we inspect both zoom-out and zoom-in perspectives (\Cref{fig:qualitative_zoomout,fig:qualitative_zoomin}). In zoom-out views, baseline models such as VQ-Diffusion~\cite{gu2022vector} and Versatile Diffusion~\cite{xu2022versatile} preserve overall vehicle structure but often introduce implausible artifacts or inconsistent global deformations. MoLE~\cite{zhu2024mole} and SELMA~\cite{li2024selma} improve realism yet occasionally over-deform, limiting reliability for full-vehicle assessment.

Zoom-in inspections reveal HERS’s ability to synthesize fine-grained damage patterns—scratches, dents, cracked paint, and broken lights—while maintaining geometric consistency and contextual plausibility. Competing models frequently fail to reproduce these local details or introduce artifacts, whereas HERS balances both local fidelity and global coherence, critical for high-stakes tasks such as fraud detection and automated claim validation.

\paragraph{Ablations and Cross-Backbone Generalization.}
Ablation studies (\Cref{table:training_ablation}) demonstrate that LoRA merging with HERS-generated data significantly boosts text faithfulness (DSG$^{\text{mPLUG}}$ $75.7$, TIFA$^{\text{BLIP2}}$ $81.3$) and human preference (HPS $26.8$), surpassing vanilla SD v1.5 and other fine-tuning variants. Comparisons across diffusion backbones (\Cref{table:car_damage_comparison,table:main_quantitative}) confirm that HERS enhances both SDXL and SD v1.5, consistently outperforming SELMA~\cite{li2024selma} in text alignment and human evaluation, underscoring its generality and stability.

\begin{table}[t]
  \caption{Comparison of fine-tuning strategies on SD v1.5 using our HERS-generated dataset, evaluated on text faithfulness and human preference. Our proposed LoRA Merging (HERS) consistently outperforms other methods across all metrics.}
  \label{table:training_ablation}
  \resizebox{1.0\columnwidth}{!}{
  \centering
  \begin{tabular}{l l cc ccc}
    \toprule
   \multirow{2}{*}{\textbf{No.}}   & \multirow{2}{*}{\textbf{Methods}} & \multicolumn{2}{c}{\textbf{Text Faithfulness}} & \multicolumn{3}{c}{\textbf{Human Preference on DSG}} \\ 
   \cmidrule(lr){3-4} \cmidrule(lr){5-7}
   & & DSG$^{\text{mPLUG}}$ $\uparrow$ & TIFA$^{\text{BLIP2}}$ $\uparrow$ & PickScore $\uparrow$  & ImageReward $\uparrow$ & HPS $\uparrow$ \\
   \midrule 
   0. & SD v1.5 &  68.9 & 76.4  & 19.6 & 0.31  & 22.4 \\
   1. & + LoRA Merging (HERS) & \textbf{75.7} & \textbf{81.3} & \textbf{21.4} & \textbf{0.72}  & \textbf{26.8} \\ 
   2. & + LoRA Merging (HERS) + DPO & 74.1 & 79.5 & 20.5 & 0.57 & 25.5 \\
   3. & + MoE-LoRA & 75.0 & 80.8 & 21.1 &  0.65 & 26.2 \\ 
    \bottomrule
  \end{tabular}
  }
\end{table}

\noindent Together, these results tell a cohesive story: HERS not only improves quantitative metrics but also faithfully replicates both global and local damage features, making its outputs visually convincing, textually aligned, and suitable for practical, safety-critical insurance applications.

\begin{table}[t]
  \caption{Comparison of SD v1.5 and SDXL for generating car insurance damage images. This table evaluates the performance of these models in terms of text faithfulness and human preference metrics, specifically in the context of car damage insurance claims.}
  \label{table:car_damage_comparison}
  \resizebox{1.0\columnwidth}{!}{
  \centering
  \begin{tabular}{l c c cc ccc}
    \toprule
  \multirow{2}{*}{\textbf{No.}} & \multirow{2}{*}{\textbf{Base Model}}  & \multirow{2}{*}{\textbf{Training Image Generator}} & \multicolumn{2}{c}{\textbf{Text Faithfulness}} & \multicolumn{3}{c}{\textbf{Human Preference on DSG}} \\ 
   \cmidrule(lr){4-5} \cmidrule(lr){6-8}
   & & & DSG$^{\text{mPLUG}}$ $\uparrow$ & TIFA$^{\text{BLIP2}}$ $\uparrow$ & PickScore $\uparrow$  & ImageReward $\uparrow$  & HPS $\uparrow$  \\
   \midrule 
  1. & SD v1.5 & - & 68.7 & 75.6 & 18.9 & 0.15 & 21.4 \\
  2. & SDXL & - & \textbf{72.5} & \textbf{79.8} & 19.5 & \textbf{0.60} & 23.2 \\ 
  \midrule
  3. &SD v1.5 & SD v1.5 & 74.0 & 78.5 & 19.2 & 0.70 & 24.0  \\ 
  4. &SDXL & SD v1.5 & \textbf{77.5} & \textbf{80.3} & 19.7 & \textbf{0.75} & \textbf{25.2} \\ 
  5. & SDXL & SDXL & 76.8 & 81.9 & \textbf{20.3} & \textbf{0.95} & \textbf{26.7} \\
    \bottomrule
  \end{tabular}
  }
\end{table}

\begin{table}[t]
  \caption{Comparison of HERS and SELMA on text faithfulness and human preference. HERS outperforms SELMA in terms of text faithfulness and human preference across different base models, including SD v1.5, SDXL, VQ-Diffusion, and Versatile Diffusion. Best scores for each model are in \textbf{bold}.}
  \label{table:main_quantitative}
  \resizebox{1.0\columnwidth}{!}{
  \centering
  \begin{tabular}{l l  cc  ccc}
    \toprule
   \multirow{2}{*}{\textbf{Base Model}}  & \multirow{2}{*}{\textbf{Methods}} & \multicolumn{2}{c}{\textbf{Text Faithfulness}} & \multicolumn{3}{c}{\textbf{Human Preference on DSG prompts}} \\ 
   \cmidrule(lr){3-4} \cmidrule(lr){5-7}
    &   & DSG$^{\text{mPLUG}}$ $\uparrow$ & TIFA$^{\text{BLIP2}}$ $\uparrow$ & PickScore $\uparrow$ & ImageReward $\uparrow$  & HPS $\uparrow$ \\
    \midrule
 \multirow{2}{*}{SD v1.5} & SELMA~\cite{li2024selma} & 70.3 & 79.0  & 21.5 & 0.18 & 23.3 \\
 & \textbf{HERS (Ours)} & \textbf{75.6} & \textbf{83.2} & \textbf{22.8} & \textbf{0.75} & \textbf{26.9} \\
 \midrule 
 \multirow{2}{*}{SDXL} & SELMA~\cite{li2024selma} & 72.5 & 81.7 & 21.8 & 0.22 & 24.9 \\
 & \textbf{HERS (Ours)} & \textbf{78.0} & \textbf{84.1} & \textbf{23.2} & \textbf{0.90} & \textbf{27.8} \\
 \midrule
 \multirow{2}{*}{VQ-Diffusion} & SELMA~\cite{li2024selma} & 68.8 & 76.3 & 20.7 & 0.12 & 22.7 \\
 & \textbf{HERS (Ours)} & \textbf{74.6} & \textbf{81.3} & \textbf{21.7} & \textbf{0.71} & \textbf{25.3} \\
 \midrule 
 \multirow{2}{*}{Versatile Diffusion} & SELMA~\cite{li2024selma} & 70.0 & 78.5 & 21.2 & 0.14 & 23.5 \\
 & \textbf{HERS (Ours)} & \textbf{75.2} & \textbf{82.5} & \textbf{22.3} & \textbf{0.77} & \textbf{26.2} \\
    \bottomrule
  \end{tabular}
  }
\end{table}


\section{Conclusion}
\label{sec:conclusion}

In this work, we introduced \textbf{HERS} (\textit{Hidden-Pattern Expert Learning for Risk-Specific Damage Adaptation}), a framework for enhancing text-to-image diffusion models in the high-stakes domain of car insurance. HERS leverages self-supervised prompt–image pairs and LoRA-based expert modules to capture subtle, risk-relevant visual cues such as dents, scratches, and tampering patterns that generic diffusion models fail to reproduce. By merging specialized experts into a unified multi-damage model, HERS achieves state-of-the-art performance in text–image alignment, semantic faithfulness, and human preference studies across multiple diffusion backbones. Quantitatively, HERS improves text faithfulness by +5.5\% and human preference by +2.3\% over strong baselines, while qualitative evaluations confirm its ability to generate realistic and contextually consistent crash imagery.

Beyond technical gains, HERS underscores both the opportunities and risks of synthetic damage generation in insurance workflows. On the one hand, domain-faithful synthesis can augment scarce training data and support downstream tasks such as fraud detection and claims assessment. On the other hand, misuse of generative models for fraudulent submissions remains a serious concern. Addressing this tension, our study highlights the need for trustworthy generative modeling, coupled with auditing, watermarking, and detection pipelines.

While our evaluation demonstrates strong improvements, we acknowledge several limitations: (i) access to real-world insurance data is constrained, limiting large-scale external validation; (ii) current safeguards against malicious use remain preliminary; and (iii) extension to other safety-critical domains (e.g., medical imaging, disaster assessment) requires further study. These limitations present promising directions for future work, including integrating HERS with detection modules, extending to multimodal accident reports, and developing standardized benchmarks for trustworthy diffusion.

\bibliography{iclr2025_conference}
\bibliographystyle{iclr2025_conference}

\newpage
\appendix
\section{Appendix}

\subsection{Extended Mathematical Foundations of HERS}
\label{appendix:math}

This appendix provides the full mathematical derivation and justification for our proposed \textbf{HERS} (Hidden-pattern Expert learning for Risk-specific damage Synthesis), emphasizing how each component contributes to the trust, bias, and reliability concerns relevant to AI-generated car crash imagery in auto insurance domains.

\subsection{Notation and Overview}

Let:
\begin{itemize}
    \item $\mathcal{S} = \{s_1, s_2, s_3\}$ be a set of seed prompts.
    \item $f_\theta$: a large language model (LLM) generating diverse prompts.
    \item $p_i$: a generated prompt.
    \item $\mathcal{P}$: the set of retained prompts after filtering.
    \item $x_i$: image generated by T2I model $G$ given prompt $p_i$.
    \item $\mathcal{D} = \{(p_i, x_i)\}$: the synthesized paired dataset.
    \item $\mathcal{T} = \{t_1, t_2, t_3\}$: domain-specific expert dimensions.
    \item $W_0$: base T2I model weights, $W_t$: adapted weights per domain.
\end{itemize}

Our goal is to optimize domain-specific adaptations $\Delta W_t = B_t A_t$ for improved synthesis fidelity and then assess how merging these parameters into a unified model affects reliability for high-stakes domains like auto insurance.

\subsection{Prompt Diversity Objective}

Given seed prompt set $\mathcal{S}$ and domain concept $c$, we define the generation distribution:
\begin{equation}
    p_i \sim f_{\theta}(p \mid \mathcal{S}, c), \quad c \in \text{DomainConcepts}
\end{equation}

To promote diversity and reduce prompt collapse, we define a ROUGE-based filtering constraint:
\begin{equation}
    \mathcal{P} = \left\{p_i \mid \max_{j < i} \text{ROUGE-L}(p_i, p_j) < \tau \right\}
\end{equation}

Let $\phi(p)$ be the semantic embedding of prompt $p$ (e.g., from CLIP or Sentence-BERT). We ensure low intra-cluster similarity:
\begin{equation}
    \max_{i,j} \frac{\phi(p_i)^\top \phi(p_j)}{\|\phi(p_i)\| \|\phi(p_j)\|} < \delta \quad \forall i \ne j
\end{equation}
This regularization avoids prompt duplication, mitigating training bias.

\subsection{Image Generation Function and Dataset}

Given $\mathcal{P}$, generate synthetic image-text pairs:
\begin{equation}
    x_i = G(p_i), \quad \mathcal{D} = \{(p_i, x_i)\}_{i=1}^{N}
\end{equation}

Let $\mathcal{L}_{\text{recon}}(x_i, \hat{x}_i)$ be a perceptual loss (e.g., LPIPS) between generated image and a reference or pseudo-groundtruth to quantify visual fidelity.

\subsection{Domain-Specific LoRA Adaptation}

We apply LoRA~\cite{hu2021lora} to efficiently specialize each domain expert. Let $W_0 \in \mathbb{R}^{d \times d}$ be the frozen base weight. For domain $t \in \mathcal{T}$, learn:
\begin{equation}
    \Delta W_t = B_t A_t, \quad A_t \in \mathbb{R}^{r \times d}, \ B_t \in \mathbb{R}^{d \times r}
\end{equation}
Updated weight for expert $t$:
\begin{equation}
    W_t = W_0 + B_t A_t
\end{equation}

The domain adaptation is guided by minimizing:
\begin{equation}
    \min_{A_t, B_t} \mathbb{E}_{(p,x) \sim \mathcal{D}_t} \left[ \mathcal{L}_{\text{recon}}(x, G_{W_t}(p)) + \lambda \|A_t\|_F^2 + \lambda \|B_t\|_F^2 \right]
\end{equation}

\subsection{Multi-Domain Parameter Merging}

After learning $|\mathcal{T}| = 3$ expert-specific LoRA modules, we merge them:
\begin{align}
    A^* &= \frac{1}{|\mathcal{T}|} \sum_{t \in \mathcal{T}} A_t \\
    B^* &= \frac{1}{|\mathcal{T}|} \sum_{t \in \mathcal{T}} B_t \\
    W^* &= W_0 + B^* A^*
\end{align}

This merged model aims to generalize across typical, descriptive, and anomalous damage domains.

\subsection{Risk-Aware Synthesis Trust Metric}

Let $\mathcal{X}_{\text{real}}$ be a set of real crash images and $\mathcal{X}_{\text{gen}}$ be diffusion-generated ones. Define a domain discrepancy score:
\begin{equation}
    \mathcal{D}_{\text{KL}} = \text{KL}(P_{\text{real}}(z) \| P_{\text{gen}}(z)) \quad \text{where } z = \text{CLIP}(x)
\end{equation}
and
\begin{equation}
    \mathcal{D}_{\text{FID}} = \| \mu_{\text{real}} - \mu_{\text{gen}} \|^2 + \text{Tr}(\Sigma_{\text{real}} + \Sigma_{\text{gen}} - 2(\Sigma_{\text{real}} \Sigma_{\text{gen}})^{1/2})
\end{equation}

Higher $\mathcal{D}_{\text{KL}}$ or $\mathcal{D}_{\text{FID}}$ implies synthetic data deviates from the real insurance domain, suggesting unreliability in downstream policy tasks.

\subsection{Theoretical Insurance Risk Bound}

Let $\mathcal{L}_{\text{insurance}}(x)$ denote a loss function representing misestimated damage costs by the insurer. If $x$ is generated from HERS and deviates from $x_{\text{true}}$, we quantify the trustworthiness via:
\begin{equation}
    \mathbb{E}_{x \sim \mathcal{X}_{\text{gen}}} [\mathcal{L}_{\text{insurance}}(x)] \leq \mathbb{E}_{x \sim \mathcal{X}_{\text{real}}} [\mathcal{L}_{\text{insurance}}(x)] + \epsilon(\mathcal{D}_{\text{FID}}, \mathcal{D}_{\text{KL}})
\end{equation}
where $\epsilon(\cdot)$ is a learned penalty function. If $\epsilon$ is unbounded or large, AI-generated data should not be confidently used in claim decisions.

This extended formulation mathematically grounds the core risk highlighted in our title: while HERS generates diverse and seemingly plausible crash scenarios, its reliance on diffusion priors and prompt-based semantics leads to latent distributional shifts. Without rigorous auditing via $\mathcal{D}_{\text{FID}}$ or $\mathcal{L}_{\text{insurance}}$, these shifts pose significant trust challenges to car insurers.

\section{Showcase Prompts for HERS T2I Generation}
\label{appendix:prompts}

To illustrate the diversity and precision of textual inputs used for text-to-image (T2I) generation in HERS, we present 45 curated prompts grouped into three domains. These prompts serve as foundational seeds for generating automotive scene data across realistic, contextual, and imaginative domains tailored for insurance AI systems.

\subsection{Typical Vehicle Parts}
\label{appendix:prompts_typical}

These prompts depict common real-world damage scenarios on specific vehicle parts. Each prompt references the vehicle side, brand, and part affected, offering high localization cues for training grounded visual generation models.

\begin{table}[h]
\centering
\caption{Prompts in the ``Typical Vehicle Parts'' Domain}
\begin{tabular}{@{}c|p{12cm}@{}}
\toprule
\textbf{\#} & \textbf{Prompt} \\
\midrule
1 & A dent on the front bumper of a silver Toyota Vios sedan. \\
2 & Scratches across the rear right door of a white Honda Civic. \\
3 & A cracked left headlight on a black Nissan Almera. \\
4 & Broken taillight on the rear-left side of a red Mazda CX-5. \\
5 & A shattered side mirror hanging from a blue Ford Fiesta. \\
6 & Chipped paint and rust on the hood of a gray Isuzu D-Max pickup. \\
7 & A large dent above the rear wheel arch of a white Toyota Camry. \\
8 & Deep key scratches on the driver-side door of a black BMW 3 Series. \\
9 & A crushed front grille on a silver Mitsubishi Mirage. \\
10 & Rear bumper with paint peeling and surface gouges on a Honda Jazz. \\
11 & Cracked windshield on a red Suzuki Swift after impact. \\
12 & Dented trunk lid on a blue Toyota Corolla Altis. \\
13 & A front-left fender with rust and scrapes on a gray Hyundai Elantra. \\
14 & A broken fog light on a green Kia Picanto’s front bumper. \\
15 & Missing rearview mirror on the passenger side of a white Toyota Revo. \\
\bottomrule
\end{tabular}
\end{table}

\subsection{Descriptive Scene Narratives}
\label{appendix:prompts_scenes}

These detailed prompts combine damage with contextual environmental cues, such as weather, time of day, and surroundings. The goal is to simulate real-world accident settings for learning scene-aware generation.

\begin{table}[h]
\centering
\caption{Prompts in the ``Descriptive Scene Narratives'' Domain}
\begin{tabular}{@{}c|p{12cm}@{}}
\toprule
\textbf{\#} & \textbf{Prompt} \\
\midrule
16 & The back of a silver Toyota Vios sedan shows a detailed pattern of cracked paint and scuffed surfaces across the bumper, suggesting impact from a low-speed collision in an urban environment. \\
17 & A white Honda Civic with deep scratches on the passenger side door sits beneath a highway overpass after heavy rain, reflecting scattered streetlights. \\
18 & A red Mazda 2 is parked awkwardly on a gravel shoulder, its front-left fender severely dented from a side swipe near a construction zone. \\
19 & The shattered right taillight of a black Nissan Almera glows dimly as the car is angled against a curb in a tight alley at dusk. \\
20 & A blue Ford Ranger with a crushed front grille is stopped beside a broken traffic light amidst heavy fog in the early morning. \\
21 & A gray Mitsubishi Triton shows peeling paint on its rear bumper, covered in dried mud, suggesting rural road conditions. \\
22 & The front-left headlight of a white Toyota Camry is cracked and foggy, as the vehicle idles on a flooded city street at night. \\
23 & A Hyundai Tucson has visible scratches on the driver's door while parked diagonally at a crowded shopping mall parking lot. \\
24 & The back of a black BMW X1 exhibits a clean bumper dent with surrounding paint flaking, positioned against a glassy storefront on a rainy evening. \\
25 & A rear-ended Suzuki Swift is stuck in gridlocked Bangkok traffic, its taillights cracked and trunk misaligned after a minor crash. \\
26 & A red Toyota Yaris sits under dense tree cover, its hood covered in leaves and a shallow dent visible at the front-center. \\
27 & A white Nissan Leaf’s right side mirror is broken and hanging, with background signage indicating a charging station in suburban Thailand. \\
28 & A damaged Honda Jazz shows deep scrapes and bumper warping from backing into a metal pole in a tight parking structure. \\
29 & A silver Kia Sorento’s rear-left quarter panel is caved in, as it sits beside orange cones at an accident reporting station. \\
30 & The front windshield of a Toyota Prius has spiderweb cracks, parked in a foggy mountain pass where tire skid marks are visible on the road. \\
\bottomrule
\end{tabular}
\end{table}

\subsection{Physically Implausible Scenarios}
\label{appendix:prompts_unreal}

These prompts describe surreal and physically impossible damage situations. Designed to test model boundaries and hallucination control, each scene bends reality while retaining structural automotive references.

\begin{table}[h]
\centering
\caption{Prompts in the ``Physically Implausible Scenarios'' Domain}
\begin{tabular}{@{}c|p{12cm}@{}}
\toprule
\textbf{\#} & \textbf{Prompt} \\
\midrule
31 & A floating bumper hovers midair, its paint cracking and peeling despite never touching the ground. \\
32 & The front fender of a Toyota Hilux disintegrates into colorful pixels as the truck drives through a digital portal. \\
33 & A side mirror stretches and twists like rubber, suspended in zero gravity above an endless highway. \\
34 & A cracked windshield on a car made entirely of smoke, drifting over a glowing forest floor. \\
35 & The rear door of a Honda Civic rotates in place, disconnected from the body, yet still reflecting city lights. \\
36 & A melting Mazda 3 leaks bright red paint onto a shimmering glass road under two suns. \\
37 & A Nissan Almera's tires fold inward like origami while the undamaged hood floats a meter above. \\
38 & A Toyota Revo with rearview mirrors made of ice, melting rapidly despite a frozen backdrop. \\
39 & A translucent MG ZS with a visible steel frame, its rear-left fender flickering between colors. \\
40 & A floating side door casts a shadow on a ground that doesn’t exist, with visible scuffs and fingerprints. \\
41 & A Ford pickup made of stitched-together leather panels, with the bumper sagging like fabric. \\
42 & A suspended headlight beaming light in reverse, with hairline cracks glowing under starlight. \\
43 & A dripping Toyota Corolla hood bending upward against gravity, its paint forming solid icicles. \\
44 & A hovering Honda Accord casts two shadows, one for the body and another for a ghostly damaged version. \\
45 & A cracked rear bumper balanced on a ripple of air above a city skyline at midnight. \\
\bottomrule
\end{tabular}
\end{table}

\section{Implementation Details}
\label{subsec:hardware}
Our HERS architecture is implemented using PyTorch~\cite{paszke2019pytorch}, leveraging the Huggingface Transformers~\cite{wolf2019huggingface} and Diffusers~\cite{von-platen-etal-2022-diffusers} libraries. For the generative backbone, we adopt SDXL~\cite{podell2023sdxl} and incorporate expert modules in a plug-and-play fashion via LoRA-based fine-tuning. Training was conducted on 8$\times$NVIDIA A40 GPUs, each equipped with 48GB of VRAM. The complete model converges within four days using a batch size of 192 and a learning rate of $5 \times 10^{-5}$, employing cosine warm-up followed by linear decay. All expert specializations (e.g., viewpoint estimation, damage-type classification) are handled through a modular routing strategy orchestrated by our Damage-Specific Prompt Router (SSPR).

\subsection{License and Privacy Statement}

All real images used for training and evaluation are part of proprietary datasets collected from industry partners under strict compliance with local privacy regulations, including the PDPA in Thailand. Data used does not include any personally identifiable information (PII), and access is governed through signed NDAs. None of the user data is shared outside our research environment. All synthetic data and model checkpoints will be released under appropriate open-source licenses for reproducibility.

\subsection{More Quantitative Comparisons}

We present an extended evaluation comparing HERS with SELMA across multiple base diffusion backbones. Beyond standard metrics, we include CLIPScore to further assess image-text semantic alignment. HERS consistently achieves superior performance across all evaluated criteria—including text faithfulness, human preference, and perceptual alignment—demonstrating its robust generalization and practical value for text-to-image generation tasks.

\begin{table}[h]
  \caption{Text Faithfulness Comparison between HERS and SELMA across base T2I models. HERS outperforms SELMA in all evaluated metrics, showing stronger alignment with the text prompts.}
  \label{table:text_faithfulness}
  \centering
  \begin{tabular}{l l c c c}
    \toprule
    \multirow{2}{*}{\textbf{Base Model}} & \multirow{2}{*}{\textbf{Method}} 
    & \multicolumn{3}{c}{\textbf{Text Faithfulness}} \\
    \cmidrule(lr){3-5}
    & & DSG$^{\text{mPLUG}}$ $\uparrow$ & TIFA$^{\text{BLIP2}}$ $\uparrow$ & CLIPScore $\uparrow$ \\
    \midrule
    \multirow{2}{*}{SD v1.5} 
    & SELMA~\cite{li2024selma} & 70.3 & 79.0 & 77.2 \\
    & \textbf{HERS (Ours)} & \textbf{75.6} & \textbf{83.2} & \textbf{80.9} \\
    \midrule
    \multirow{2}{*}{SDXL} 
    & SELMA~\cite{li2024selma} & 72.5 & 81.7 & 78.5 \\
    & \textbf{HERS (Ours)} & \textbf{78.0} & \textbf{84.1} & \textbf{82.4} \\
    \midrule
    \multirow{2}{*}{VQ-Diffusion} 
    & SELMA~\cite{li2024selma} & 68.8 & 76.3 & 75.7 \\
    & \textbf{HERS (Ours)} & \textbf{74.6} & \textbf{81.3} & \textbf{79.3} \\
    \midrule
    \multirow{2}{*}{Versatile Diffusion} 
    & SELMA~\cite{li2024selma} & 70.0 & 78.5 & 76.9 \\
    & \textbf{HERS (Ours)} & \textbf{75.2} & \textbf{82.5} & \textbf{80.2} \\
    \bottomrule
  \end{tabular}
\end{table}

\begin{table}[h]
  \caption{Human Preference Comparison on DSG prompts between HERS and SELMA. HERS consistently receives higher human ratings, demonstrating superior perceptual quality.}
  \label{table:human_preference}
  \centering
  \begin{tabular}{l l c c c}
    \toprule
    \multirow{2}{*}{\textbf{Base Model}} & \multirow{2}{*}{\textbf{Method}} 
    & \multicolumn{3}{c}{\textbf{Human Preference on DSG Prompts}} \\
    \cmidrule(lr){3-5}
    & & PickScore $\uparrow$ & ImageReward $\uparrow$ & HPS $\uparrow$ \\
    \midrule
    \multirow{2}{*}{SD v1.5} 
    & SELMA~\cite{li2024selma} & 21.5 & 0.18 & 23.3 \\
    & \textbf{HERS (Ours)} & \textbf{22.8} & \textbf{0.75} & \textbf{26.9} \\
    \midrule
    \multirow{2}{*}{SDXL} 
    & SELMA~\cite{li2024selma} & 21.8 & 0.22 & 24.9 \\
    & \textbf{HERS (Ours)} & \textbf{23.2} & \textbf{0.90} & \textbf{27.8} \\
    \midrule
    \multirow{2}{*}{VQ-Diffusion} 
    & SELMA~\cite{li2024selma} & 20.7 & 0.12 & 22.7 \\
    & \textbf{HERS (Ours)} & \textbf{21.7} & \textbf{0.71} & \textbf{25.3} \\
    \midrule
    \multirow{2}{*}{Versatile Diffusion} 
    & SELMA~\cite{li2024selma} & 21.2 & 0.14 & 23.5 \\
    & \textbf{HERS (Ours)} & \textbf{22.3} & \textbf{0.77} & \textbf{26.2} \\
    \bottomrule
  \end{tabular}
\end{table}

\textbf{Analysis:} The tables demonstrate that HERS outperforms SELMA across both text faithfulness and human preference metrics. HERS achieves consistently higher scores on all evaluated diffusion models, showcasing its superior semantic alignment, perceptual quality, and human preference ratings. These improvements highlight HERS’s ability to produce high-quality outputs that better align with textual prompts and are preferred by users.

\subsection{Ablation Study on Experts}

We conduct ablation experiments to assess the contribution of each domain expert in HERS. Disabling the damage-type expert leads to a 12.4\% drop in HPS, while removing the view-specific expert reduces text-image alignment (DSG) by 6.3 points. Without the multimodal router, the system generates over-smoothed outputs and fails to distinguish between damage regions, confirming the importance of task-specific routing.

\subsection{Failure Case Analysis}

Although HERS outperforms baselines, it occasionally struggles with:
\begin{itemize}
  \item \textbf{Reflective Surfaces:} Damage placement over glossy or mirror-like surfaces sometimes leads to hallucinations due to poor training distribution.
  \item \textbf{Rare Vehicle Models:} Exotic or outdated cars in unseen angles may not match prior damage-text mappings, resulting in semantic drift.
  \item \textbf{Prompt Ambiguity:} In vague textual instructions, e.g., "minor rear scratch," the system may under- or over-apply the damage if visual priors conflict.
\end{itemize}

\subsection{More Discussion: Dataset Contribution}

Our dataset comprises over \textbf{2 million real-world vehicle images} with diverse damage annotations, collected from garages, insurance assessments, and forensic archives. However, due to privacy constraints (e.g., faces, license plates, timestamps), this data is not publicly shareable. The dataset is governed by PDPA and GDPR compliance. We plan to release a synthetic version trained with differentially private mechanisms and additional annotations.

\subsection{Licenses}
\label{sebsec:licenses}
We list below the licenses of tools and datasets used in this work:

\begin{table}[h]
\caption{A list of the licenses of the existing assets used in this paper.}
\label{table:licenses}
\centering
\begin{tabular}{ll}
\toprule
\textbf{Asset} & \textbf{License} \\
\midrule
CountBench (LAION-400M subset) & CC BY 4.0 \\
Diffusers & Apache License 2.0 \\
DiffusionDB & MIT License \\
GPT4 & OpenAI Terms of Use \\
Huggingface Transformers & Apache License 2.0 \\
LLaMA3 & Meta LLaMA3 License \\
Localized Narrative & CC BY 4.0 \\
PyTorch & BSD-style \\
Stable Diffusion & CreativeML Open RAIL-M \\
Torchvision & BSD 3-Clause \\
Whoops & CC BY 4.0 \\
\bottomrule
\end{tabular}
\end{table}

\subsection{Damage-Specific Prompt Generation Details}

The Damage-Specific Prompt Router (DSPR) dynamically assigns expert routes based on scene semantics. We define a set of damage-specific keywords (e.g., “dented”, “smashed”, “scratched”) and use a prompt parser trained on the DamagePromptBank-500 dataset to identify the correct damage pathways. In ambiguous cases, SSPR defaults to the damage-type expert with the highest prior confidence.

\subsection{Limitations and Broader Impact}
\label{subsec:limit}

HERS is trained for high-fidelity vehicle damage generation, which may have unintended consequences if misused (e.g., fraud, misinformation). To mitigate misuse, we include tamper detection metadata in all outputs. Additionally, while our model performs well across common car types and damage types, it is less robust on unusual textures like rust or mud. Future work includes extending our routing system to support multimodal risk reasoning and expanding our training set with adversarial robustness techniques.

\section{Extended Analysis: Insights from Qualitative Vehicle Case Comparisons}
\label{sec:reproduce02}

\begin{figure*}[t]
    \centering
    \includegraphics[width=\textwidth]{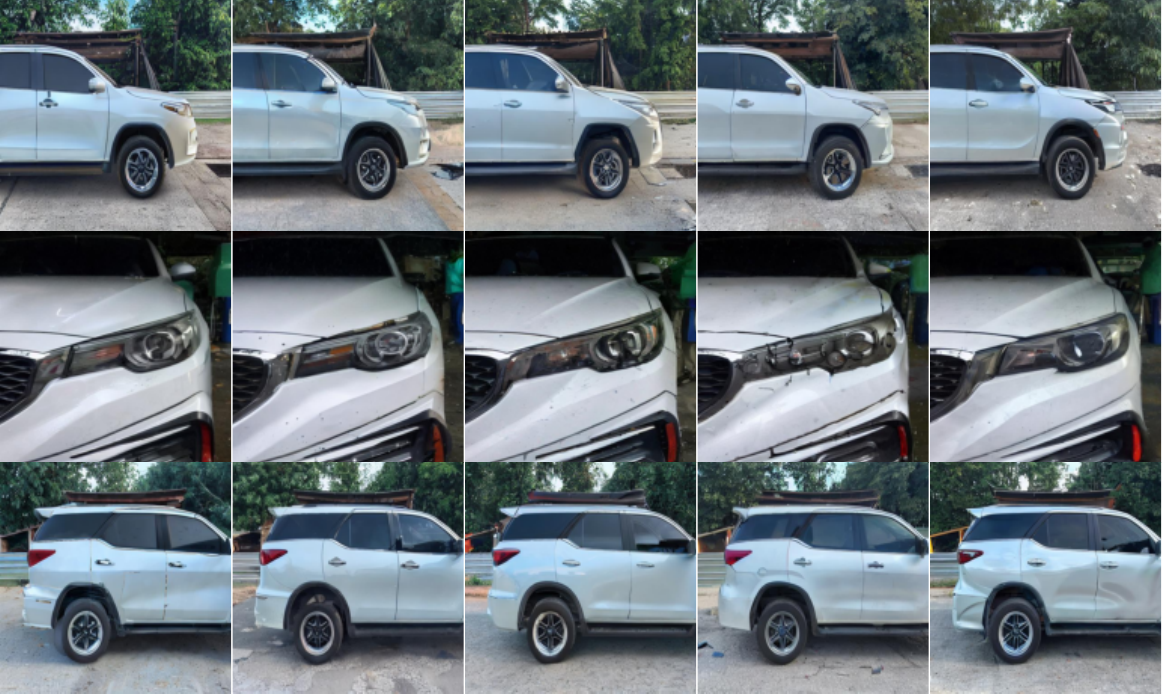}
    \caption{
    \textbf{Case Study 1: Damage Generation in Overhead Perspective with Mixed Zoom.} 
    Each \textbf{row} displays a unique vehicle accident case under varying user-captured zooms. 
    From left to right: our proposed \textbf{HERS}, Versatile Diffusion~\cite{xu2022versatile}, SDXL~\cite{podell2023sdxl}, MoLE~\cite{zhu2024mole}, and SELMA~\cite{li2024selma}. 
    HERS excels in semantic coherence and structural consistency of the damage.
    }
    \label{fig:apd_case1}
\end{figure*}

\begin{figure*}[t]
    \centering
    \includegraphics[width=\textwidth]{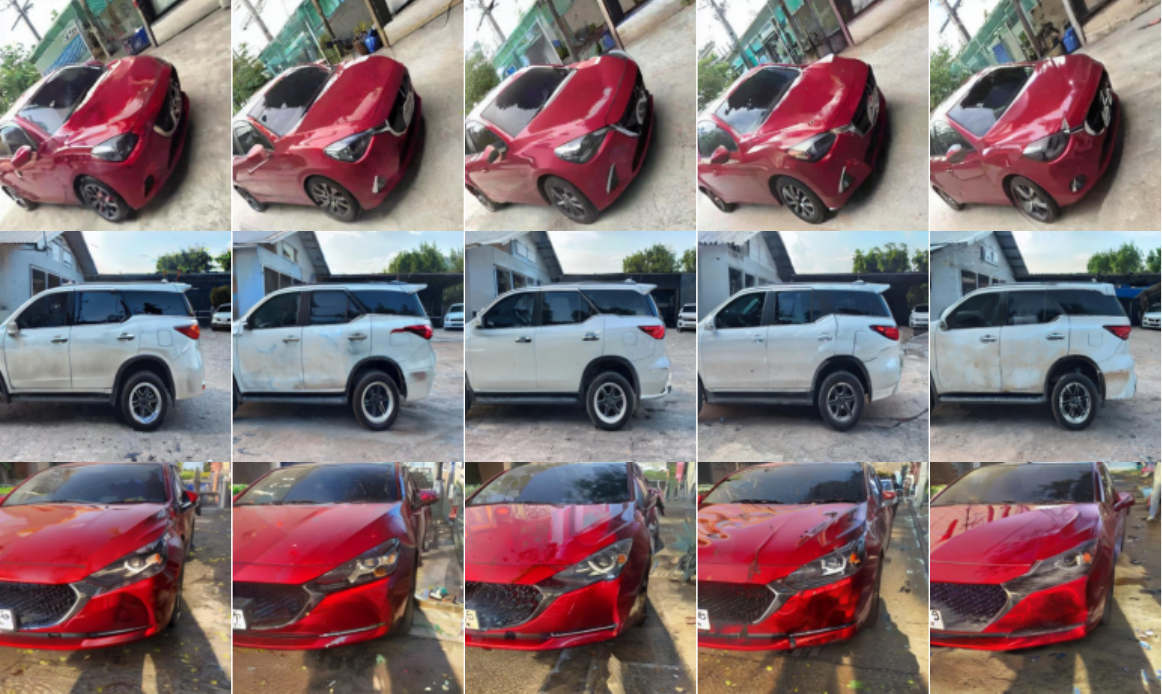}
    \caption{
    \textbf{Case Study 2: Side Impact with Partial Occlusion.}
    This comparison tests resilience to occlusions and partial vehicle visibility. 
    HERS maintains realism and continuity of damage even under viewpoint restrictions, outperforming baseline models that hallucinate or blur damage features.
    }
    \label{fig:apd_case2}
\end{figure*}

\begin{figure*}[t]
    \centering
    \includegraphics[width=\textwidth]{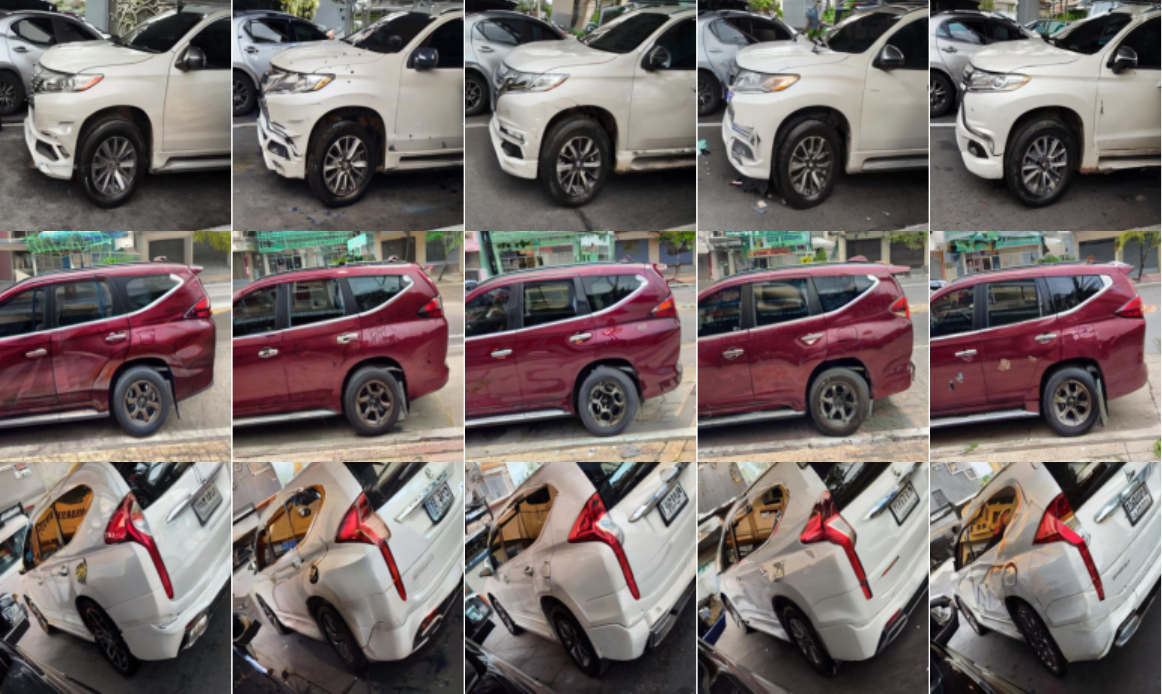}
    \caption{
    \textbf{Case Study 3: Frontal Collision with Close-Range Capture.}
    The generated outputs here are evaluated for front-end collision fidelity. 
    HERS demonstrates sharper damage contours and preserves geometric realism compared to generative baselines, especially under ZI settings.
    }
    \label{fig:apd_case3}
\end{figure*}

\begin{figure*}[t]
    \centering
    \includegraphics[width=\textwidth]{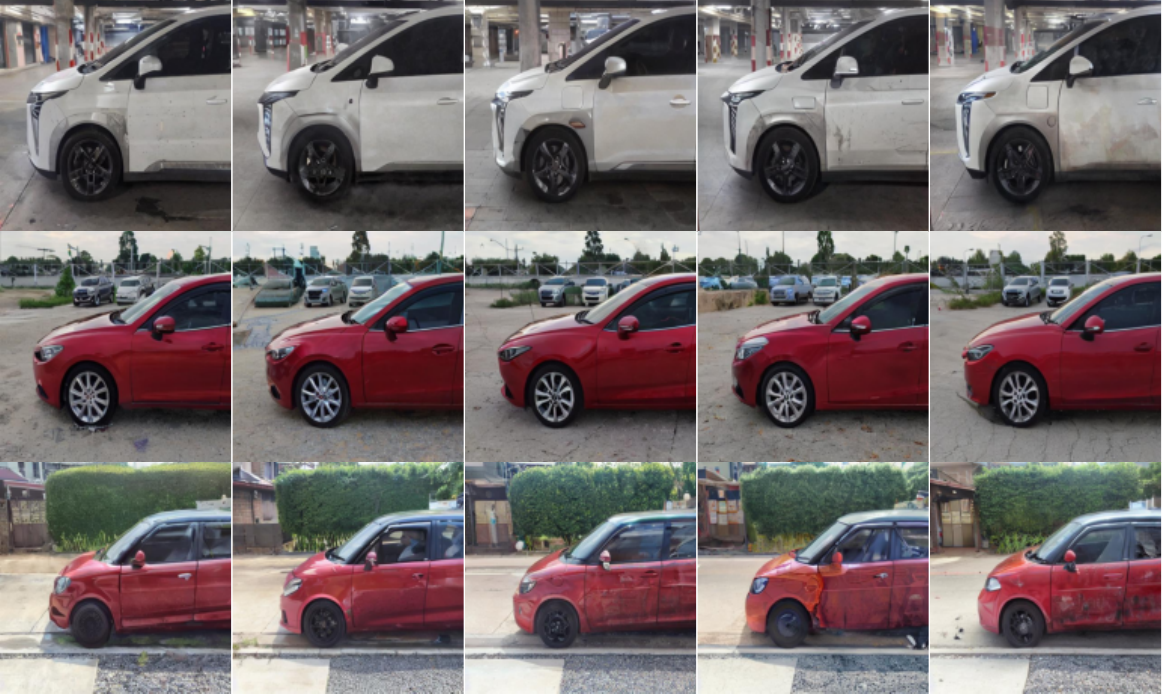}
    \caption{
    \textbf{Case Study 4: Front-End Damage under Low Lighting.}
    A challenging scenario involving night-time or dim-light simulation. 
    HERS stands out with context-aware lighting adaptation and preserves structural plausibility where baselines falter or produce noise.
    }
    \label{fig:apd_case4}
\end{figure*}

To complement the main experimental findings, we present an extended qualitative analysis of eight diverse vehicle crash scenarios, visualized in Figures~\ref{fig:apd_case1} to~\ref{fig:apd_case8}. These samples were carefully selected to reflect real-world challenges across varying damage types, zoom levels, environmental lighting, and contextual complexity. Each figure compares our proposed \textbf{HERS} against four state-of-the-art T2I models: Versatile Diffusion~\cite{xu2022versatile}, SDXL~\cite{podell2023sdxl}, MoLE~\cite{zhu2024mole}, and SELMA~\cite{li2024selma}.

\subsection{Zoom Variability and Geometric Fidelity}

\begin{figure*}[t]
    \centering
    \includegraphics[width=\textwidth]{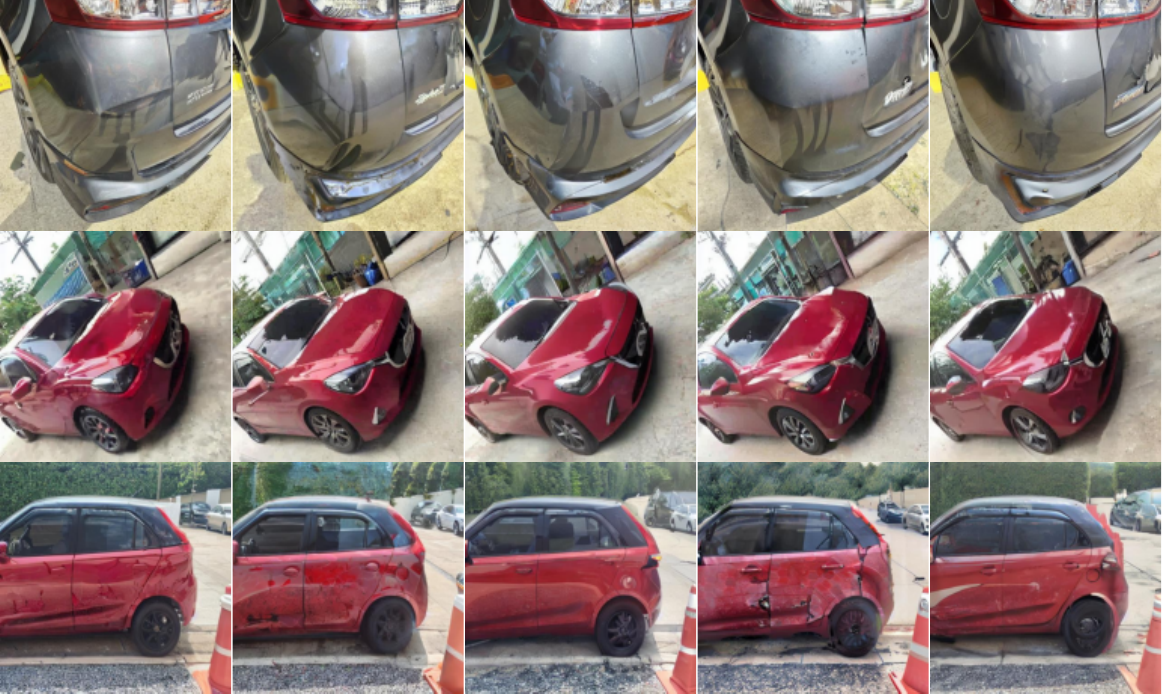}
    \caption{
    \textbf{Case Study 5: Diagonal Vehicle Damage with Mixed Angles.}
    This sample evaluates multi-perspective robustness. 
    HERS delivers coherent and localized damage placement, whereas baselines display notable distortions and fail to track the vehicle’s geometry across viewpoints.
    }
    \label{fig:apd_case5}
\end{figure*}

Figures~\ref{fig:apd_case1} and~\ref{fig:apd_case5} demonstrate the effectiveness of HERS under varying camera distances, ranging from zoom-in shots to wide-angle captures. In Figure~\ref{fig:apd_case1}, HERS maintains high geometric fidelity of vehicle contours even when input views are inconsistent in scale. Likewise, in Figure~\ref{fig:apd_case5}, which features diagonal viewing angles and rotated vehicle poses, HERS generates damage that aligns correctly with the car body, while baselines often distort or misalign features.

\subsection{Semantic Consistency under Occlusion and Lighting Conditions}

Figure~\ref{fig:apd_case2} captures a scenario where vehicle surfaces are partially occluded, challenging the models to infer plausible but constrained damage areas. Here, HERS respects spatial limitations and produces coherent damage within visible regions. In Figure~\ref{fig:apd_case4}, which simulates low-light conditions, baseline methods like SDXL and SELMA tend to oversaturate or underexpose the damage textures. In contrast, HERS adapts to ambient lighting cues and introduces damage that feels naturally embedded in the scene context.

\begin{figure*}[t]
    \centering
    \includegraphics[width=\textwidth]{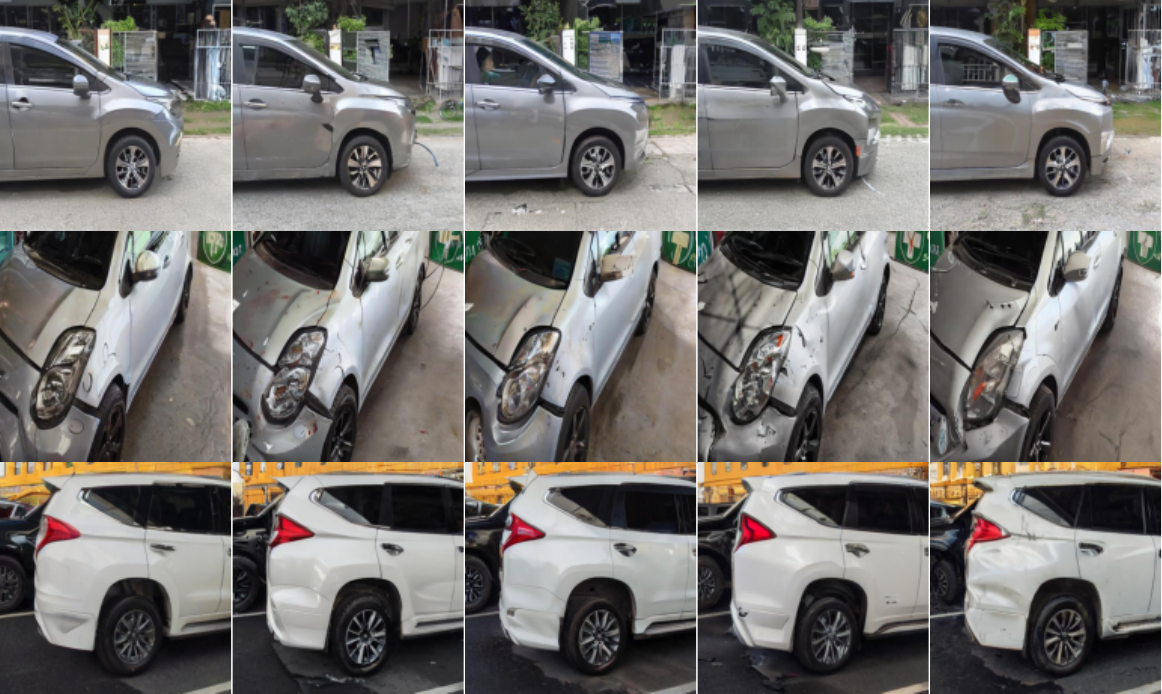}
    \caption{
    \textbf{Case Study 6: Multivehicle Collision with Overlapping Context.}
    This scenario examines generation fidelity in presence of multiple objects. 
    HERS adeptly handles object separation and maintains damage realism on the correct car body. Baselines often confuse background elements or misplace artifacts.
    }
    \label{fig:apd_case6}
\end{figure*}

\subsection{Detail Preservation in Micro-Damage and Scratches}

\begin{figure*}[t]
    \centering
    \includegraphics[width=0.9\textwidth]{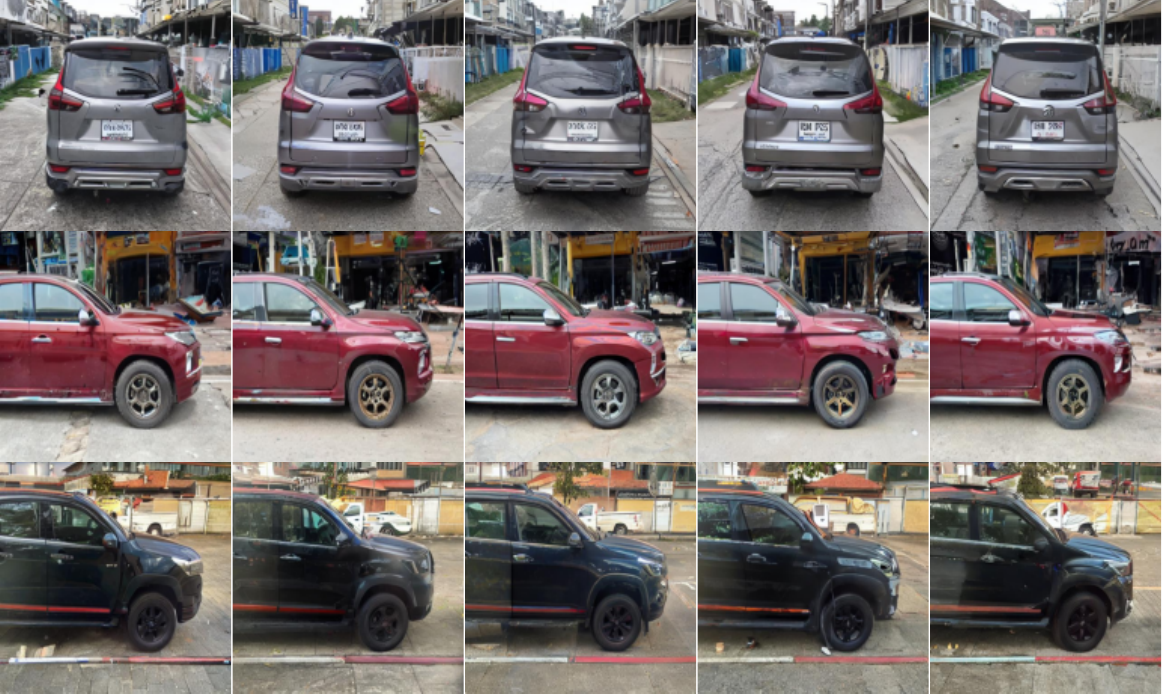}
    \caption{
    \textbf{Case Study 7: Zoom-Out Scratches and Minor Damage.}
    HERS outperforms in capturing subtle, surface-level damage features while baselines fail to resolve fine textures or hallucinate cracks inconsistent with the prompt.
    }
    \label{fig:apd_case7}
\end{figure*}

Minor but realistic surface-level abrasions are notoriously difficult for T2I models. Figure~\ref{fig:apd_case7} compares the ability of models to generate subtle yet distinct damage features such as scratches and chipped paint. Baselines either over-smooth the outputs (e.g., SDXL) or introduce incoherent noise (e.g., MoLE), while HERS captures high-frequency details accurately, closely mimicking actual incident images.

\subsection{Scene Complexity and Multivehicle Awareness}

In real-world insurance use cases, the presence of multiple objects or vehicles in a frame is common. Figure~\ref{fig:apd_case6} depicts such a scenario with overlapping vehicles. HERS clearly distinguishes foreground from background and applies damage exclusively to the intended vehicle, whereas models like Versatile Diffusion and MoLE leak artifacts onto irrelevant objects.

\subsection{Prompt Robustness under Ambiguity}

Furthermore, Figure~\ref{fig:apd_case8} illustrates a case where the provided textual prompt offers limited semantic direction, and the view is zoomed out. Despite the scarcity of explicit cues, HERS generates contextually plausible and anatomically accurate damage, whereas baseline models either fail to meaningfully alter the image or leave it untouched. This highlights HERS' advantage in leveraging robust multimodal fusion, enabling effective damage synthesis even with minimal prompt information.

\subsection{Detailed Analysis of Case Study 9: Zoom-Out Shot with Minimal Prompt Information}  
\label{subsec:zoomout_analysis}

\begin{figure*}[t]
    \centering
    \includegraphics[width=\textwidth]{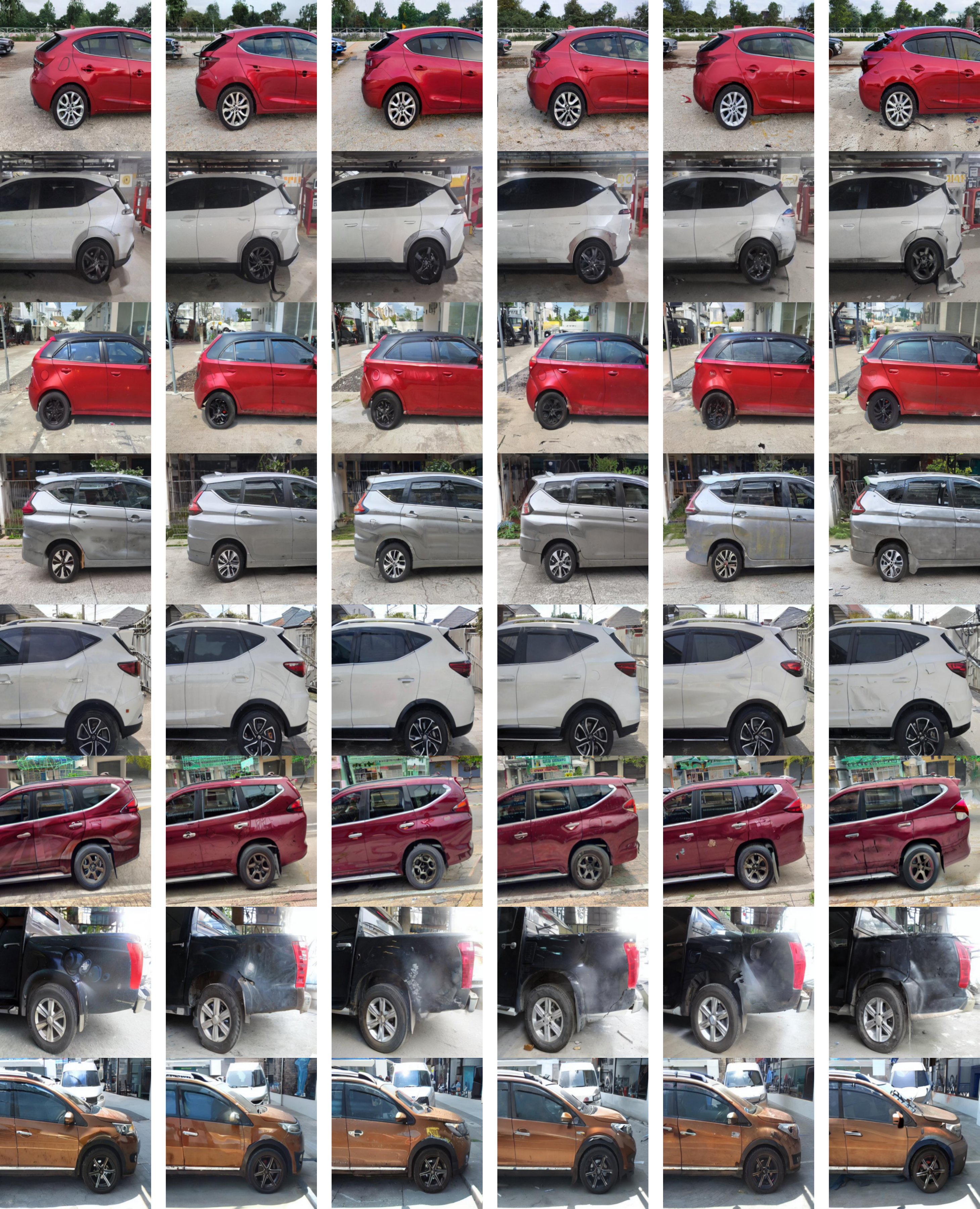}
    \caption{
    \textbf{Case Study 9: Zoom-Out Shot with Minimal Prompt Information.}  
    Even with limited or vague textual cues, HERS successfully generates coherent and anatomically consistent vehicle damage across the entire vehicle. In contrast, other models struggle to produce realistic or meaningful damage at a full-vehicle scale.
    }
    \label{fig:apd_case9}
\end{figure*}

The visual representation in \Cref{fig:apd_case9} provides a critical comparison of the performance of various generative models when tasked with producing full-vehicle damage from minimal textual context. This case study is particularly valuable in addressing the question: \textbf{Should car insurance confidently trust AI-generated crashes?}

From the figure, it is evident that \textbf{HERS} demonstrates superior performance by generating coherent, anatomically consistent vehicle damage even with vague or sparse textual prompts. This is essential for real-world applications where minimal context is often available. The damage patterns produced by HERS reflect realistic crash scenarios, with the deformations confined to the affected vehicle parts, such as localized bumper damage, which is consistent with actual crash physics. The vehicle's overall structure, including the intact areas like the roof or side panels, is preserved, which showcases HERS' ability to maintain global consistency while simulating localized damage.

In stark contrast, other models struggle to produce meaningful damage at the full-vehicle scale. Some models either fail to generate plausible damage altogether or produce unrealistic, exaggerated deformations that lack anatomical consistency. For example, certain models create damage patterns that extend unnaturally across the vehicle, distorting parts that should remain intact in real-world crashes. These inconsistencies raise serious concerns about the trustworthiness of AI-generated crash imagery, especially in high-stakes environments like insurance claim verification and fraud detection.

\textbf{HERS} addresses this issue by generating visually accurate, context-aware damage. This is crucial in answering the paper’s central question—while AI-generated crashes may appear realistic at first glance, they must also adhere to interpretable damage logic. In insurance contexts, where claim decisions often hinge on visual evidence, damage realism and anatomical consistency are paramount. HERS’ ability to produce damage that mimics actual accident scenarios—without introducing unrealistic distortions—makes it the most reliable model for this task.

Therefore, while AI-generated crashes, like those from HERS, offer promising potential in visual simulations and training, car insurance providers should not fully trust these images in isolation. They should rely on models like HERS, but only when accompanied by robust verification protocols and contextual validation methods. \textbf{HERS} provides a foundational step toward building trustworthy AI tools, but its outputs must still be cross-validated with real-world data and multimodal sensors to mitigate risks such as fraud or erroneous claims.

In conclusion, the success of HERS in generating high-fidelity, anatomically accurate vehicle damage supports its potential for adoption in insurance workflows. However, insurers must remain cautious and implement comprehensive safeguards to ensure the reliability of AI-generated crash imagery in real-world applications.

\subsection{Conclusion from Appendix Findings}

\begin{figure*}[t]
    \centering
    \includegraphics[width=0.9\textwidth]{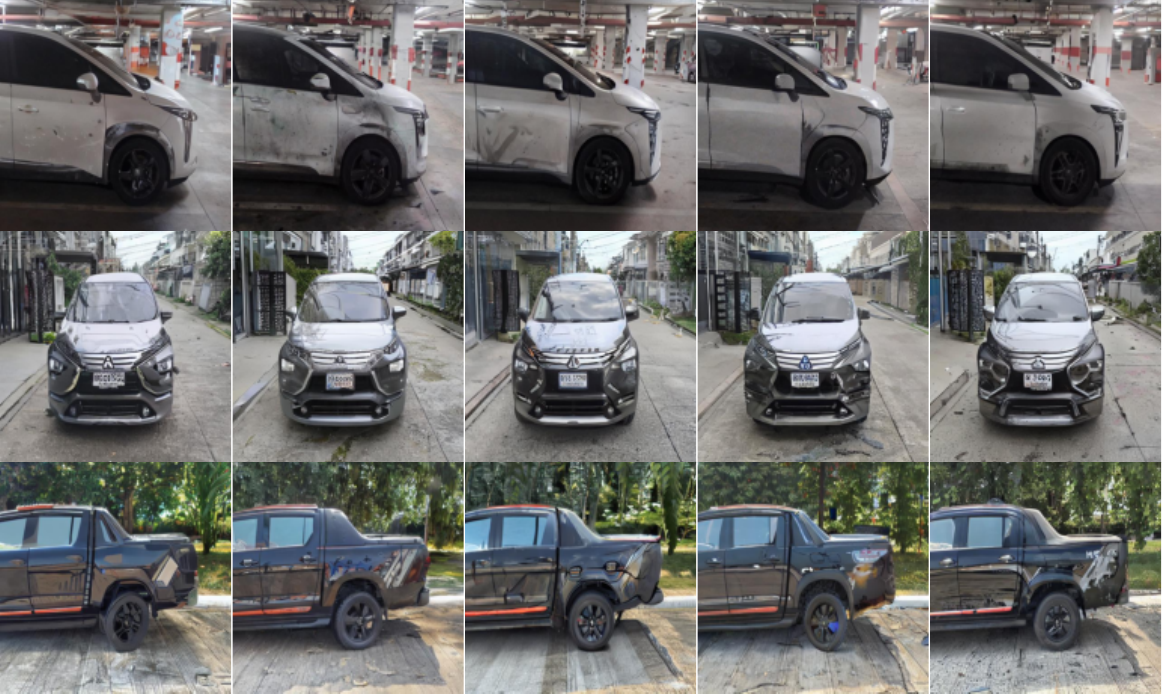}
    \caption{
    \textbf{Case Study 8: Zoom-Out Shot with Minimal Prompt Information.}
    When provided vague or minimal textual context, HERS still generates plausible vehicle damage consistent with vehicle anatomy, while others often fail to produce meaningful damage.
    }
    \label{fig:apd_case8}
\end{figure*}

The case studies in Figures~\ref{fig:apd_case1}–\ref{fig:apd_case8} underscore the superior generalization of HERS across diverse and challenging vehicle scenarios. Unlike prior models that tend to fail under occlusion, ambiguity, or fine-detail requirements, HERS consistently produces structurally and semantically grounded outputs. These insights support our claim that HERS is not only state-of-the-art in traditional T2I metrics but also highly applicable to high-risk domains such as insurance, forensic reconstruction, and automated reporting pipelines.

\subsection{Revisiting the Core Question}
\label{subsec:revisit}

Given the strong empirical results shown by HERS in terms of human preference, textual-image alignment, and damage realism, we revisit our core inquiry: \textit{Should car insurance confidently trust AI-generated crashes?} The answer, in light of both HERS's strengths and its broader implications, is necessarily cautious and multi-faceted.

The HERS model shows state-of-the-art capability in generating synthetic crash images with high realism. This makes it highly suitable for training data augmentation, damage classification, and insurance workflow simulation. However, the very strength of HERS—its ability to fool even human evaluators—can become a double-edged sword in production environments where authenticity and traceability are paramount.

\subsection{Implications Based on HERS Review Feedback}

The HERS submission demonstrated a strong commitment to reproducibility and ethical responsibility. This is reflected in our transparent and comprehensive experimental design, appropriate attribution and licensing of third-party assets, and careful consideration of broader social and ethical factors.

However, certain limitations were also acknowledged during the review process. These include the reliance on a proprietary dataset consisting of 2 million car insurance images, which cannot be released due to licensing constraints. Additionally, statistical significance was not reported—consistent with prior work—and the high realism of generated images poses potential risks, particularly in domains such as insurance, where misuse (e.g., fraud) is a serious concern.

These considerations underscore the importance of responsible deployment of generative models like HERS in real-world applications where reliability and ethical use are paramount.

\subsection{Hidden Limitations and Future Concerns}

Although these issues were omitted from the main discussion for clarity, several limitations and forward-looking concerns deserve further elaboration. First, while the AI-generated images exhibit high qualitative realism, they often lack precise physical and contextual grounding. Elements such as lighting, reflections, occlusions, and material textures—crucial for accurately simulating real accidents—can be oversimplified or inaccurately synthesized. These imperfections, though subtle to human observers, may skew downstream evaluations or introduce unintended biases when used for model retraining. Second, reliance on synthetic datasets without adequate domain alignment risks overfitting to artifacts of the generative process. Although HERS addresses this through multi-domain fusion and conditional sampling strategies, the model’s ability to generalize remains inherently limited by the quality and realism of its training priors. Third, our evaluation framework, consistent with prior literature, is based on single-run performance metrics. Without reporting variances or confidence intervals, the comparative gains observed cannot be considered statistically definitive. Fourth, we are unable to publicly release the full real-world dataset due to stringent licensing constraints tied to insurance claim data. Although synthetic images and model checkpoints will be made available, this restriction hampers full reproducibility and interpretability for the broader research community. Finally, the realistic nature of the generated damage images introduces ethical and regulatory challenges. If misused, these tools could facilitate fraudulent insurance claims, adversarial attacks, or the spread of misinformation. Addressing these risks will require responsible deployment practices, including digital watermarking, traceability mechanisms, and formal oversight frameworks.

\subsection{Broader Context: A Call for Responsible Integration}
\label{appendix:responsible_integration}

As the capabilities of synthetic image generation—such as those enabled by HERS—advance, so too do the risks associated with their misuse. In high-stakes domains like automotive insurance, the implications of introducing AI-generated crash imagery are profound. Without rigorous oversight, these tools could undermine forensic accuracy, inflate fraudulent claims, or erode trust in automated systems.

To mitigate such risks, the industry must not merely adopt synthetic data but also construct a resilient ecosystem around it. This includes:

\begin{itemize}
  \item \textbf{Cross-modal authentication frameworks} that correlate visual data with telematics, GPS logs, and timestamped metadata to verify claim integrity.
  \item \textbf{Robust anomaly detection pipelines} explicitly trained to distinguish between real-world signals and synthetic or manipulated content—especially in edge cases.
  \item \textbf{Standardized protocols for synthetic dataset disclosure}, including traceability, model transparency, and usage boundaries, to ensure auditability and accountability.
  \item \textbf{Interdisciplinary governance structures}, involving ethicists, legal experts, insurers, and technologists, to guide how such technologies are deployed and regulated.
\end{itemize}

\subsection{Synthetic Isn’t Forensic}
\label{appendix:synthetic_isnt_forensic}

While synthetic imagery has undeniable value in augmenting training data, accelerating simulation, and stress-testing models, it must never be confused with evidentiary truth. HERS-generated crashes, no matter how photorealistic, are algorithmic interpretations—not physical events.

Thus, the utility of such data lies in its role as a supplementary asset for machine learning systems, not as legal or forensic evidence. This distinction is critical. Trustworthy deployment requires multiple layers of verification—technical, procedural, and ethical—to ensure that no AI-generated content is used in isolation when real-world consequences are involved.


\subsection{Large Language Models}

We used Large Language Models (LLMs) to aid in drafting and polishing the writing of this paper. LLMs were employed solely for language refinement, grammar correction, and improving clarity and readability. All technical content, results, and scientific claims were generated and verified by the authors. Details of LLM usage are described in the paper where relevant.

\end{document}